\pdfoutput=1
\documentclass[conference]{IEEEtran}
\usepackage{subfigure}
\usepackage{url}
\usepackage{balance}
\usepackage{booktabs}
\usepackage{tikz}
\usepackage{epstopdf}
\usepackage{graphicx}
\usepackage{bm}
\usepackage{amsmath,amssymb}

\usepackage{algorithm}
\usepackage{algorithmic}

\newcommand{\gscore}[0]{gSide}
\newcommand{\galgo}[0]{gMSV}
\newcommand{\gssc}[0]{gSSC}
\newcommand{\duga}[0]{Conf}
\newcommand{\dugb}[0]{Ratio}
\newcommand{\dugc}[0]{Gtest}
\newcommand{\dugd}[0]{HSIC}
\newcommand{\topk}[0]{Freq}
\newcommand{\side}[0]{MSV}

\newcommand{\neu}[0]{\emph{neuropsychological tests}}
\newcommand{\flo}[0]{\emph{flow cytometry}}
\newcommand{\pla}[0]{\emph{plasma luminex}}
\newcommand{\fre}[0]{\emph{freesurfer}}
\newcommand{\ave}[0]{\emph{overall brain microstructure}}
\newcommand{\dti}[0]{\emph{localized brain microstructure}}
\newcommand{\seg}[0]{\emph{brain volumetry}}

\newtheorem{Theorem}{\textsc{Theorem}}
\newtheorem{Definition}{\textsc{Definition}}

\begin{document}
%----------------------------------------------------------------------------------------
\title{Mining Brain Networks using Multiple Side Views for Neurological Disorder Identification}
\author{\IEEEauthorblockN{
Bokai Cao\IEEEauthorrefmark{1},
Xiangnan Kong\IEEEauthorrefmark{2},
Jingyuan Zhang\IEEEauthorrefmark{1},
Philip S. Yu\IEEEauthorrefmark{1}\IEEEauthorrefmark{3} and
Ann B. Ragin\IEEEauthorrefmark{4}}
\IEEEauthorblockA{\IEEEauthorrefmark{1}Department of Computer Science, University of Illinois at Chicago, IL, USA; \{caobokai, jzhan8, psyu\}@uic.edu}
\IEEEauthorblockA{\IEEEauthorrefmark{2}Department of Computer Science, Worcester Polytechnic Institute, MA, USA; xkong@wpi.edu}
\IEEEauthorblockA{\IEEEauthorrefmark{3}Institute for Data Science, Tsinghua University, Beijing, China}
\IEEEauthorblockA{\IEEEauthorrefmark{4}Department of Radiology, Northwestern University, IL, USA; ann-ragin@northwestern.edu}}

\maketitle

%-----------------------------------------------
\begin{abstract}
Mining discriminative subgraph patterns from graph data has attracted great interest in recent years. It has a wide variety of applications in disease diagnosis, neuroimaging, \emph{etc.} Most research on subgraph mining focuses on the graph representation alone. However, in many real-world applications, the side information is available along with the graph data. For example, for neurological disorder identification, in addition to the brain networks derived from neuroimaging data, hundreds of clinical, immunologic, serologic and cognitive measures may also be documented for each subject. These measures compose multiple side views encoding a tremendous amount of supplemental information for diagnostic purposes, yet are often ignored. In this paper, we study the problem of discriminative subgraph selection using multiple side views and propose a novel solution to find an optimal set of subgraph features for graph classification by exploring a plurality of side views. We derive a feature evaluation criterion, named {\gscore}, to estimate the usefulness of subgraph patterns based upon side views. Then we develop a branch-and-bound algorithm, called {\galgo}, to efficiently search for optimal subgraph features by integrating the subgraph mining process and the procedure of discriminative feature selection. Empirical studies on graph classification tasks for neurological disorders using brain networks demonstrate that subgraph patterns selected by the multi-side-view guided subgraph selection approach can effectively boost graph classification performances and are relevant to disease diagnosis.
\end{abstract}

%\category{H.2.8}{Database Management}{Database Applications-Data Mining}
%\category{I.5.2}{Pattern Recognition}{Design Methodology-Feature Evaluation and Selection}
%\terms{Algorithm, Experimentation, Performance}
%\keywords{Subgraph Pattern, Graph Mining, Side Information, Brain Network}

%-----------------------------------------------
\section{Introduction}
\label{sec:intro}
Recent years have witnessed an increasing amount of data in the form of graph representations, which involve complex structures, \emph{e.g.}, brain networks, social networks. These data are inherently represented as a set of nodes and links, instead of feature vectors as traditional data. For example, brain networks are composed of brain regions as the nodes, \emph{e.g.}, \emph{insula}, \emph{hippocampus}, \emph{thalamus}, and functional/structural connectivities between the brain regions as the links. The linkage structure in these brain networks can encode tremendous information about the mental health of the human subjects. For example, in the brain networks derived from functional magnetic resonance imaging (fMRI), functional connections/links can encode the correlations between the functional activities of brain regions. While structural links in diffusion tensor imaging (DTI) brain networks can capture the number of neural fibers connecting different brain regions. The complex structures and the lack of vector representations within these graph data raise a challenge for data mining. An effective model for mining the graph data should be able to extract a set of subgraph patterns for further analysis. Motivated by such challenges, graph mining research problems, in particular graph classification, have received considerable attention in the last decade.

The graph classification problem has been studied extensively. Conventional approaches focus on mining discriminative subgraphs from graph view alone. This is usually feasible for applications like molecular graph analysis, where a large set of graph instances with labels are available. For brain network analysis, however, usually we only have a small number of graph instances, ranging from 30 to 100 brain networks \cite{kong2013discriminative}. In these applications, the information from the graph view alone are usually not sufficient for mining important subgraphs. We notice that, fortunately, the side information is available along with the graph data for neurological disorder identification. For example, in neurological studies, hundreds of clinical, immunologic, serologic and cognitive measures may be available for each subject \cite{cao2014tensor,cao2015determinants}, in addition to brain networks derived from the neuroimaging data, as shown in Figure~\ref{fig:sideview}. These measures compose multiple side views which contain a tremendous amount of supplemental information for diagnostic purposes. It is desirable to extract valuable information from a plurality of side views to guide the process of subgraph mining in brain networks.

\begin{figure}[t]
\centering
    \begin{minipage}[l]{\columnwidth}
      \centering
      \includegraphics[width=1\textwidth]{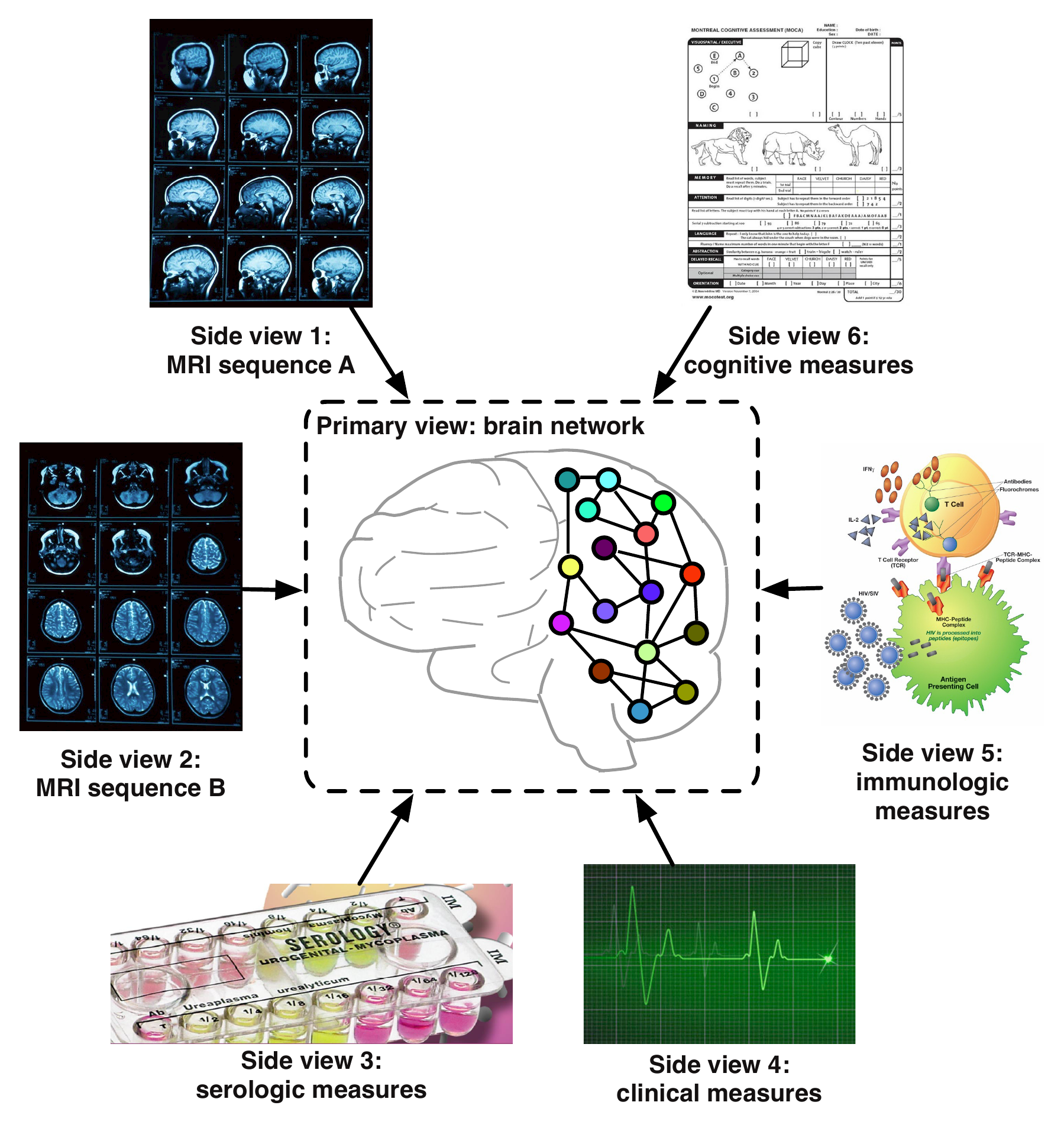}
    \end{minipage}
  \caption{An example of multiple side views associated with brain networks in medical studies.}\label{fig:sideview}%\vspace{-15pt}
\end{figure}

Despite its value and significance, the feature selection problem for graph data using auxiliary views has not been studied in this context so far. There are two major difficulties in learning from multiple side views for graph classification, as follows:

\noindent\textbf{The primary view in graph representation}: Graph data naturally composes the primary view for graph mining problems, from which we want to select discriminative subgraph patterns for graph classification. However, it raises a challenge for data mining with the complex structures and the lack of vector representations. Conventional feature selection approaches in vector spaces usually assume that a set of features are given before conducting feature selection. In the context of graph data, however, subgraph features are embedded within the graph structures and usually it is not feasible to enumerate the full set of subgraph features for a graph dataset before feature selection. Actually, the number of subgraph features grows exponentially with the size of graphs.

\noindent\textbf{The side views in vector representations}: In many applications, side information is available along with the graph data and usually exists in the form of vector representations. That is to say, an instance is represented by a graph and additional vector-based features at the same time. It introduces us to the problem of how to leverage the relationship between the primary graph view and a plurality of side views, and how to facilitate the subgraph mining procedure by exploring the vector-based auxiliary views. For example, in brain networks, discriminative subgraph patterns for neurological disorders indicate brain injuries associated with particular regions. Such changes can potentially express in other medical tests of the subject, \emph{e.g.}, clinical, immunologic, serologic and cognitive measures. Thus, it would be desirable to select subgraph features that are consistent with these side views.

\begin{figure}[t]
\centering
  \subfigure[Conventional methods treat side views and subgraph patterns separately and may only concatenate them in the final step for graph classification.]{\label{fig:method1}
    \begin{minipage}[l]{\columnwidth}
      \centering
      \includegraphics[width=1\textwidth]{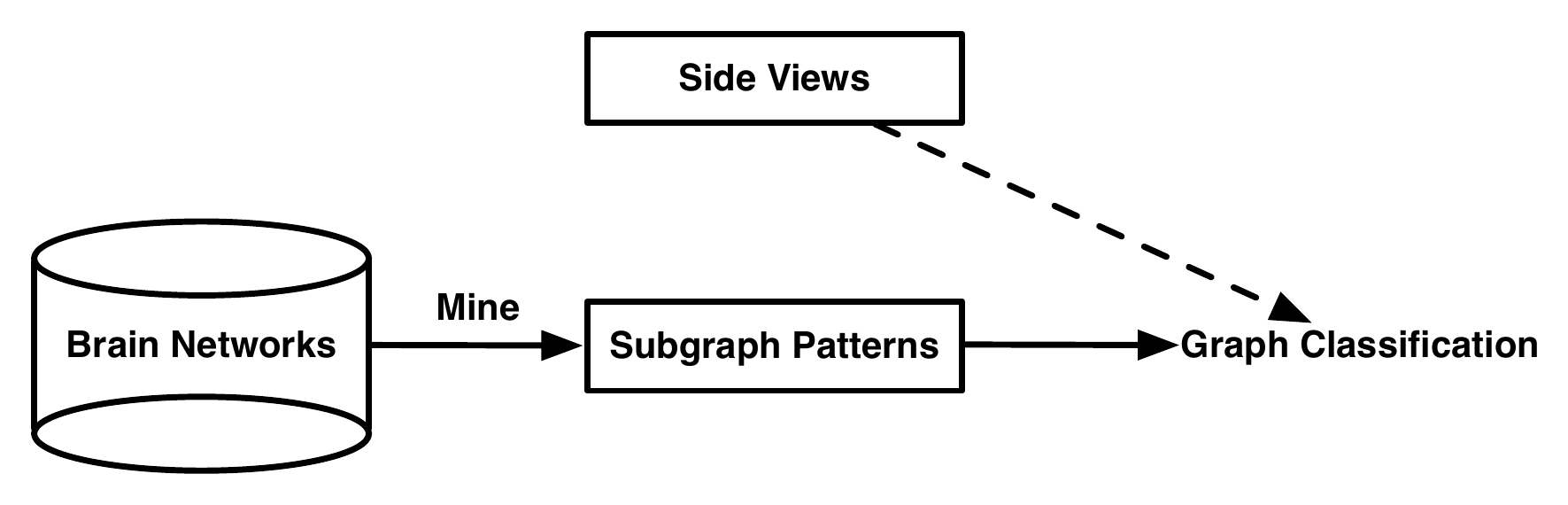}
    \end{minipage}
  }
\subfigure[Our method uses side views as guidance for the process of selecting subgraph patterns.]{\label{fig:method2}
    \begin{minipage}[l]{\columnwidth}
      \centering
      \includegraphics[width=1\textwidth]{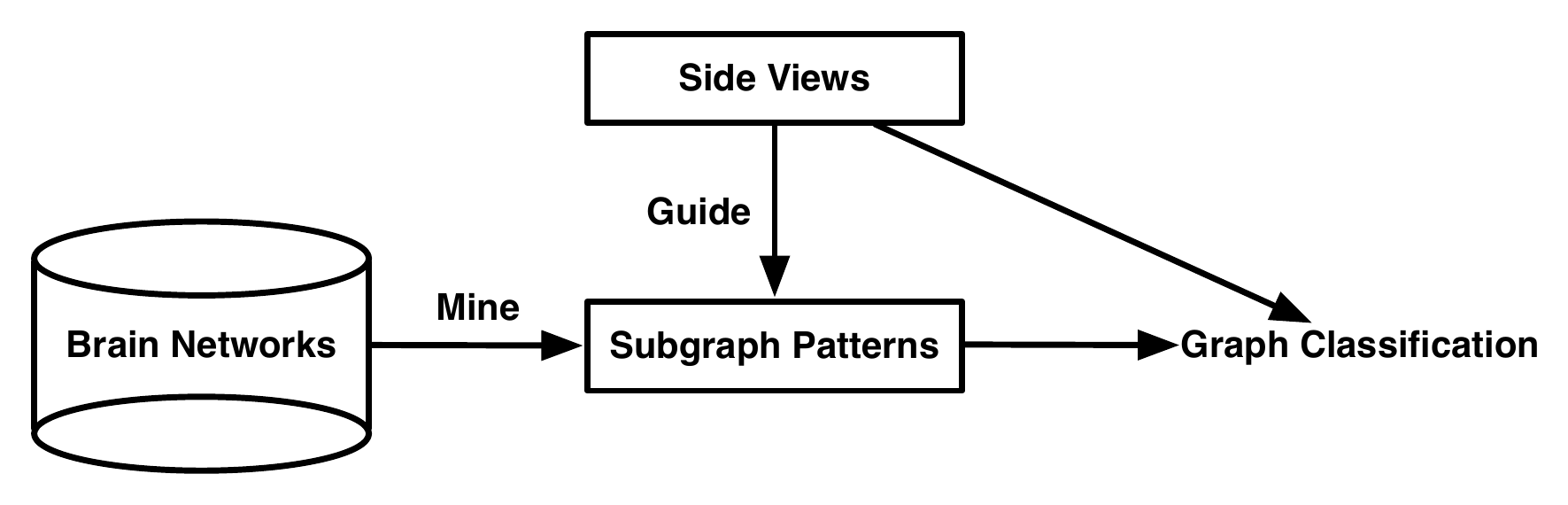}
    \end{minipage}
  }
\caption{Two strategies of leveraging side views in feature selection process for graph classification.}\label{fig:method}
\end{figure}

Figure~\ref{fig:method1} illustrates the process of selecting subgraph patterns in conventional graph classification approaches. Obviously, the valuable information embedded in side views is not fully leveraged in feature selection process. Most subgraph mining approaches focus on the drug discovery problem which have access to a great amount of graph data for chemical compounds. For neurological disorder identification, however, there are usually limited subjects with a small sample size of brain networks available. Therefore, it is critical to learn knowledge from other possible sources. We notice that transfer learning can borrow supervision knowledge from the source domain to help the learning on the target domain, \emph{e.g.}, finding a good feature representation \cite{dai2007co}, mapping relational knowledge \cite{mihalkova2007mapping,mihalkova2009transfer}, and learning across graph database \cite{shi2012transfer}. However, as far as we know, they do not look at transferring complementary information from vector-based side views to graph database whose instances are complex structural graphs.

To solve the above problems, in this paper, we introduce a novel framework for discriminative subgraph selection using multiple side views. Our framework is illustrated in Figure~\ref{fig:method2}. In contrast to existing subgraph mining approaches that focus on a single view of the graph representation, our method can explore multiple vector-based side views to find an optimal set of subgraph features for graph classification. We first verify side information consistency via statistical hypothesis testing. Based on auxiliary views and the available label information, we design an evaluation criterion for subgraph features, named {\gscore}. By deriving a lower bound, we develop a branch-and-bound algorithm, called {\galgo}, to efficiently search for optimal subgraph features with pruning, thereby avoiding exhaustive enumeration of all subgraph features. In order to evaluate our proposed model, we conduct experiments on graph classification tasks for neurological disorders, using fMRI and DTI brain networks. The experiments demonstrate that our subgraph selection approach using multiple side views can effectively boost graph classification performances. Moreover, we show that {\galgo} is more efficient by pruning the subgraph search space via {\gscore}.

%The rest of the paper is organized as follows. Section~\ref{sec:relatedwork} presents a brief review on related work of feature selection for graph classification and multi-view learning. Section~\ref{sec:problem} states the problem of discriminative subgraph selection using multiple side views. Section~\ref{sec:data} describes our data collections and verifies side information consistency. Section~\ref{sec:method} presents the evaluation criterion leveraging side information, {\gscore}, and the searching strategy with pruning, {\galgo}. Section~\ref{sec:exp} evaluates our proposed method on neuroimaging datasets. We conclude the paper and give directions for future work in Section~\ref{sec:conclusion}.

%-----------------------------------------------
\section{Problem Formulation}
\label{sec:problem}

Before presenting the subgraph feature selection model, we first introduce the notations that will be used throughout this paper. Let $\mathcal{D}=\{G_1,\cdots,G_n\}$ denote the graph dataset, which consists of $n$ graph objects. The graphs within $\mathcal{D}$ are labeled by $[y_1,\cdots,y_n]^\top$, where $y_i\in\{-1,+1\}$ denotes the binary class label of $G_i$.

\begin{Definition}[Side view]
A side view is a set of vector-based features $\mathbf{z}_i=[z_1,\cdots,z_d]^\top$ associated with each graph object $G_i$, where $d$ is the dimensionality of this view. A side view is denoted as $\mathcal{Z}=\{\mathbf{z}_1,\cdots,\mathbf{z}_n\}$.
\end{Definition}

We assume that there are multiple side views $\{\mathcal{Z}^{(1)},\cdots,\mathcal{Z}^{(v)}\}$ available for the graph dataset $\mathcal{D}$, where $v$ is the number of side views. We employ kernels $\kappa^{(p)}$ on $\mathcal{Z}^{(p)}$, such that $\kappa^{(p)}_{ij}$ represents the similarity between $G_i$ and $G_j$ from the perspective of the $p$-th view.
%In this way, the data representations in side views are not limited in feature vectors. Actually, matrices and even high-order tensors can be included, as long as we can define a kernel on them.
The RBF kernel is used as the default kernel in this paper, unless otherwise specified:
\begin{equation}
\kappa^{(p)}_{ij}=\text{exp}\left(-\frac{\|\mathbf{z}_i^{(p)}-\mathbf{z}_j^{(p)}\|_2^2}{d^{(p)}}\right)
\label{eq:rbf}
\end{equation}

\begin{Definition}[Graph] A graph is represented as $G =(V,E)$, where $V=\{v_1,\cdots,v_{n_v}\}$ is the set of vertices, $E\subseteq V\times V$ is the set of edges.
\end{Definition}

\begin{Definition}[Subgraph]
Let $G'=(V',E')$ and $G=(V,E)$ be two graphs. $G'$ is a subgraph of $G$ (denoted as $G'\subseteq G$) iff $V'\subseteq V$ and $E'\subseteq E$. If $G'$ is a subgraph of $G$, then $G$ is supergraph of $G'$.
\end{Definition}

% Table:  List of Notations
\begin{table*}[t]
\centering
%\small
\caption{Important notations.}\label{tab_notation}
\begin{tabular}{ll}
\toprule
Symbol & Definition and Description\\
\midrule
$| . |$ & cardinality of a set\\
$\| . \|$ & norm of a vector\\
$\mathcal{D}=\{G_1,\cdots,G_n\}$ & given graph dataset, $G_i$ denotes the $i$-th graph in the dataset\\
$\mathbf{y}=[y_1,\cdots,y_n]^\top$ & class label vector for graphs in $\mathcal{D}$, $y_i\in\{-1,+1\}$\\
$\mathcal{S}=\{g_1,\cdots,g_m\}$ & set of all subgraph patterns in the graph dataset $\mathcal{D}$\\
$\mathbf{f}_i=[f_{i1},\cdots,f_{in}]^\top$ & binary vector for subgraph pattern $g_i$, $f_{ij}=1$ iff $g_i\subseteq G_j$, otherwise $f_{ij}=0$\\
$\mathbf{x}_j=[x_{1j},\cdots,x_{mj}]^\top$ & binary vector for $G_j$ using subgraph patterns in $\mathcal{S}$, $x_{ij}=1$ iff $g_i\subseteq G_j$, otherwise $x_{ij}=0$\\
$X=[x_{ij}]^{m\times n}$ & matrix of all binary vectors in the dataset, $X=[\mathbf{x}_1,\cdots,\mathbf{x}_n]=[\mathbf{f}_1,\cdots,\mathbf{f}_m]^\top\in\{0,1\}^{m\times n}$\\
$\mathcal{T}$ & set of selected subgraph patterns, $\mathcal{T}\subseteq\mathcal{S}$\\
$\mathcal{I}_\mathcal{T}\in\{0,1\}^{m\times m}$ & diagonal matrix indicating which subgraph patterns are selected from $\mathcal{S}$ into $\mathcal{T}$\\
\emph{min\_sup} & minimum frequency threshold; frequent subgraphs are contained by at least \emph{min\_sup}$\times|\mathcal{D}|$ graphs\\
$k$ & number of subgraph patterns to be selected\\
$\lambda^{(p)}$ & weight of the $p$-th side view (default: 1)\\
$\kappa^{(p)}$ & kernel function on the $p$-th side view (default: RBF kernel)\\
\bottomrule
\end{tabular}
%\vspace{-10pt}
\end{table*}

In this paper, we adopt the idea of subgraph-based graph classification approaches, which assume that each graph object $G_j$ is represented as a binary vector $\mathbf{x}_j=[x_{1j},\cdots,x_{mj}]^\top$ associated with the full set of subgraph patterns $\{g_1,\cdots,g_m\}$ for the graph dataset $\{G_1,\cdots,G_n\}$. Here $x_{ij}\in\{0,1\}$ is the binary feature of $G_j$ corresponding to the subgraph pattern $g_i$, and $x_{ij}=1$ iff $g_i$ is a subgraph of $G_j$ ($g_i\subseteq G_j$), otherwise $x_{ij}=0$. Let $X=[x_{ij}]^{m\times n}$ denote the matrix consisting of binary feature vectors using $\mathcal{S}$ to represent the graph dataset $\mathcal{D}$. $X=[\mathbf{x}_1,\cdots,\mathbf{x}_n]=[\mathbf{f}_1,\cdots,\mathbf{f}_m]^\top\in\{0,1\}^{m\times n}$. The full set $\mathcal{S}$ is usually too large to be enumerated. There is usually only a subset of subgraph patterns $\mathcal{T}\subseteq\mathcal{S}$ relevant to the task of graph classification. We briefly summarize the notations used in this paper in Table~\ref{tab_notation}.

The key issue of discriminative subgraph selection using multiple side views is how to find an optimal set of subgraph patterns for graph classification by exploring the auxiliary views. This is non-trivial due to the following problems:
\begin{itemize}
\item How to leverage the valuable information embedded in multiple side views to evaluate the usefulness of a set of subgraph patterns?
\item How to efficiently search for the optimal subgraph patterns without exhaustive enumeration in the primary graph space?
\end{itemize}

In the following sections, we will first introduce the optimization framework for selecting discriminative subgraph features using multiple side views. Next, we will describe our subgraph mining strategy using the evaluation criterion derived from the optimization solution.

\section{Data Analysis}
\label{sec:data}

A motivation for this work is that the side information could be strongly correlated with the health state of a subject. Before proceeding, we first introduce real-world data used in this work and investigate whether the available information from side views has any potential impact on neurological disorder identification.

\subsection{Data Collections}
\label{sec:dataset}

In this paper, we study the real-world datasets collected from the Chicago Early HIV Infection Study at Northwestern University \cite{ragin2012structural}. The clinical cohort includes 56 HIV (positive) and 21 seronegative controls (negative). The datasets contain functional magnetic resonance imaging (fMRI) and diffusion tensor imaging (DTI) for each subject, from which brain networks can be constructed, respectively.

For fMRI data, we used DPARSF toolbox\footnote{\url{http://rfmri.org/DPARSF}} to extract a sequence of responds from each of the 116 anatomical volumes of interest (AVOI), where each AVOI represents a different brain region. The correlations of brain activities among different brain regions are computed. Positive correlations are used as links among brain regions. For details, functional images were realigned to the first volume, slice timing corrected, and normalized to the MNI template and spatially smoothed with an 8-mm Gaussian kernel. The linear trend of time series and temporally band-pass filtering (0.01-0.08 Hz) were removed. Before the correlation analysis, several sources of spurious variance were also removed from the data through linear regression: (i) six parameters obtained by rigid body correction of head motion, (ii) the whole-brain signal averaged over a fixed region in atlas space, (iii) signal from a ventricular region of interest, and (iv) signal from a region centered in the white matter. Each brain is represented as a graph with 90 nodes corresponding to 90 cerebral regions, excluding 26 cerebellar regions.

For DTI data, we used FSL toolbox\footnote{\url{http://fsl.fmrib.ox.ac.uk/fsl/fslwiki}} to extract the brain networks. The processing pipeline consists of the following steps: (i) correct the distortions induced by eddy currents in the gradient coils and use affine registration to a reference volume for head motion, (ii) delete non-brain tissue from the image of the whole head \cite{smith2002fast,jenkinson2005bet2}, (iii) fit the diffusion tensor model at each voxel, (iv) build up distributions on diffusion parameters at each voxel, and (v) repetitively sample from the distributions of voxel-wise principal diffusion directions. As with the fMRI data, the DTI images were parcellated into 90 regions (45 for each hemisphere) by propagating the Automated Anatomical Labeling (AAL) to each image \cite{tzourio2002automated}. Min-max normalization was applied on link weights.

In addition, for each subject, hundreds of clinical, imaging, immunologic, serologic and cognitive measures were documented. Seven groups of measurements were investigated in our datasets, including {\neu}, {\flo}, {\pla}, {\fre}, {\ave}, {\dti}, {\seg}. Each group can be regarded as a distinct view that partially reflects subject status, and measurements from different medical examinations can provide complementary information. Moreover, we preprocessed the features by min-max normalization before employing the RBF kernel on each view.

\subsection{Verifying Side Information Consistency}
\label{sec:ttest}

We study the potential impact of side information on selecting subgraph patterns via statistical hypothesis testing. Side information consistency suggests that the similarity of side view features between instances with the same label should have higher probability to be larger than that with different labels. We use hypothesis testing to validate whether this statement holds in the fMRI and DTI datasets.

For each side view, we first construct two vectors $\mathbf{a}_s^{(p)}$ and $\mathbf{a}_d^{(p)}$ with an equal number of elements, sampled from the sets $\mathcal{A}_s^{(p)}$ and $\mathcal{A}_d^{(p)}$, respectively:
\begin{equation}
\mathcal{A}_s^{(p)}=\{\kappa^{(p)}_{ij}|y_iy_j=1\}
\label{eq:a_s}
\end{equation}
\vspace{-15pt}
\begin{equation}
\mathcal{A}_d^{(p)}=\{\kappa^{(p)}_{ij}|y_iy_j=-1\}
\label{eq:a_d}
\end{equation}

Then, we form a two-sample one-tail t-test to validate the existence of side information consistency. We test whether there is sufficient evidence to support the hypothesis that the similarity score in $\mathbf{a}_s^{(p)}$ is larger than that in $\mathbf{a}_d^{(p)}$. The null hypothesis is $H_0: \mu_s^{(p)}-\mu_d^{(p)}\leq0$, and the alternative hypothesis is $H_1: \mu_s^{(p)}-\mu_d^{(p)}>0$, where $\mu_s^{(p)}$ and $\mu_d^{(p)}$ represent the sample means of similarity scores in the two groups, respectively.

\begin{table}[!ht]
\caption{Hypothesis testing results (p-values) to verify side information consistency.}
\small
\label{tab:result_ttest}
\centering
\begin{tabular}{lcc}
\toprule%----------------------------
Side views	&fMRI dataset &DTI dataset\\
\midrule %---------------------------
{\neu}	&1.3220e-20	&3.6015e-12 \\
{\flo}  	&5.9497e-57	&5.0346e-75 \\
{\pla}  	&9.8102e-06	&7.6090e-06 \\
{\fre}  	&2.9823e-06	&1.5116e-03 \\
{\ave}	&1.0403e-02	&8.1027e-03 \\
{\dti}	&3.1108e-04	&5.7040e-04 \\
{\seg}	&2.0024e-04	&1.2660e-02 \\
\bottomrule%-------------------------
\end{tabular}
\end{table}

The t-test results, p-values, are summarized in Table~\ref{tab:result_ttest}. The results show that there is strong evidence, with significance level $\alpha=0.05$, to reject the null hypothesis on the two datasets. In other words, we validate the existence of side information consistency in neurological disorder identification, thereby paving the way for our next study of leveraging multiple side views for discriminative subgraph selection.

\section{Multi-Side-View Discriminative \\Subgraph Selection}
\label{sec:method}

In this section, we address the first problem discussed in Section~\ref{sec:problem} by formulating the discriminative subgraph selection problem as a general optimization framework as follows:
\begin{equation}
\mathcal{T}^*=\operatornamewithlimits{argmin}_{\mathcal{T}\subseteq\mathcal{S}}\mathcal{F}(\mathcal{T})~~~\text{s.t.}~|\mathcal{T}|\le k
\label{eq:argmin1}
\end{equation}
where $|\cdot|$ denotes the cardinality and $k$ is the maximum number of feature selected. $\mathcal{F}(\mathcal{T})$ is the evaluation criterion to estimate the score (can be the lower the better in this paper) of a subset of subgraph patterns $\mathcal{T}$. $\mathcal{T}^*$ denotes the optimal set of subgraph patterns $\mathcal{T}^*\subseteq\mathcal{S}$.

\subsection{Exploring Multiple Side Views: \gscore}

Following the observations in Section~\ref{sec:ttest} that the side view information is clearly correlated with the prespecified label information, we assume that the set of optimal subgraph patterns should have the following properties. The similarity/distance between instances in the space of subgraph features should be consistent with that in the space of a side view. That is to say, if two instances are similar in the space of the $p$-th view (\emph{i.e.}, a high $\kappa^{(p)}_{ij}$ value), they should also be close to each other in the space of subgraph features (\emph{i.e.}, a small distance between subgraph feature vectors). On the other hand, if two instances are dissimilar in the space of the $p$-th view (\emph{i.e.}, a low $\kappa^{(p)}_{ij}$ value), they should be far away from each other in the space of subgraph features (\emph{i.e.}, a large distance between subgraph feature vectors). Therefore, our objective function could be to minimize the distance between subgraph features of each pair of similar instances in each side view, and maximize the distance between dissimilar instances. This idea is formulated as follows:
\begin{equation}
\operatornamewithlimits{argmin}_{\mathcal{T}\subseteq\mathcal{S}}\frac{1}{2}\sum_{p=1}^v\lambda^{(p)}\sum_{i,j=1}^n
\|\mathcal{I}_\mathcal{T}\mathbf{x}_i-\mathcal{I}_\mathcal{T}\mathbf{x}_j\|^2_2\Theta^{(p)}_{ij}
\label{eq:J1}
\end{equation}
where $\mathcal{I}_\mathcal{T}$ is a diagonal matrix indicating which subgraph features are selected into $\mathcal{T}$ from $\mathcal{S}$, $(\mathcal{I}_\mathcal{T})_{ii}=1$ iff $g_i\in\mathcal{T}$, otherwise $(\mathcal{I}_\mathcal{T})_{ii}=0$. The parameters $\lambda^{(p)}\ge0$ are employed to control the contributions from each view.% and $\sum_{p=1}^v\lambda^{(p)}=1$.
\begin{align}\label{eq:theta}
\Theta_{ij}^{(p)}
&=\left\{
\begin{array}{ll}
    \frac{1}{|\mathcal{H}^{(p)}|}~&~(i,j)\in\mathcal{H}^{(p)}\\
    -\frac{1}{|\mathcal{L}^{(p)}|}~&~(i,j)\in\mathcal{L}^{(p)}
\end{array}
\right.
\end{align}
where $\mathcal{H}^{(p)}=\{(i,j)|\kappa^{(p)}_{ij}\ge\mu^{(p)}\}$, $\mathcal{L}^{(p)}=\{(i,j)|\kappa^{(p)}_{ij}<\mu^{(p)}\}$, and $\mu^{(p)}$ is the mean value of $\kappa^{(p)}_{ij}$, \emph{i.e.}, $\frac{1}{n^2}\sum_{i,j=1}^n\kappa^{(p)}_{ij}$. This normalization is to balance the effect of similar instances and dissimilar instances.

Intuitively, Eq.~(\ref{eq:J1}) will minimize the distance between subgraph features of similar instance-pairs with $\kappa^{(p)}_{ij}\ge\mu^{(p)}$, while maximize the distance between dissimilar instance-pairs with $\kappa^{(p)}_{ij}<\mu^{(p)}$ in each view. In this way, the side view information is effectively used to guide the process of discriminative subgraph selection. The fact verified in Section~\ref{sec:ttest} that the side view information is clearly correlated with the prespecified label information can be very useful, especially in the semi-supervised setting.

With prespecified information for labeled graphs, we further consider that the optimal set of subgraph patterns should satisfy the following constraints: labeled graphs in the same class should be close to each other; labeled graphs in different classes should be far away from each other. Intuitively, these constraints tend to select the most discriminative subgraph patterns based on the graph labels. Such an idea has been well explored in the context of dimensionality reduction and feature selection \cite{bar2005learning,tang2006pairwise}.% and deep neural networks \cite{chopra2005learning,zhang2014supervised}.

The constraints above can be mathematically formulated as minimizing the loss function:
\begin{equation}
\operatornamewithlimits{argmin}_{\mathcal{T}\subseteq\mathcal{S}}\frac{1}{2}\sum_{i,j=1}^n\|\mathcal{I}_\mathcal{T}\mathbf{x}_i-\mathcal{I}_\mathcal{T}\mathbf{x}_j\|^2_2\Omega_{ij}
\label{eq:J2}
\end{equation}
where
\begin{align}\label{eq:omega}
\Omega_{ij}
&=\left\{
\begin{array}{ll}
    \frac{1}{|\mathcal{M}|}~&~(i,j)\in\mathcal{M}\\
    -\frac{1}{|\mathcal{C}|}~&~(i,j)\in\mathcal{C}\\
    0~&~\text{otherwise}
\end{array}
\right.
\end{align}
and $\mathcal{M}=\{(i,j)|y_i y_j=1\}$ denotes the set of pairwise constraints between graphs with the same label, and $\mathcal{C}=\{(i,j)|y_i y_j=-1\}$ denotes the set of pairwise constraints between graphs with different labels.

By defining matrix $\Phi\in\mathbb{R}^{n\times n}$ as
\begin{equation}
\Phi_{ij}=\Omega_{ij}+\sum_{p=1}^v\lambda^{(p)}\Theta^{(p)}_{ij}
\label{eq:W}
\end{equation}
we can combine and rewrite the function in Eq.~(\ref{eq:J1}) and Eq.~(\ref{eq:J2}) as
\begin{equation}
\begin{split}
\mathcal{F}(\mathcal{T})&=\frac{1}{2}\sum_{i=1}^n\sum_{j=1}^n
    \|\mathcal{I}_\mathcal{T}\mathbf{x}_i-\mathcal{I}_\mathcal{T}\mathbf{x}_j\|^2_2\Phi_{ij}\\
&=\text{tr}(\mathcal{I}^{\top}_\mathcal{T}X(D-\Phi)X^{\top}\mathcal{I}_\mathcal{T})\\
&=\text{tr}(\mathcal{I}^{\top}_\mathcal{T}XLX^{\top}\mathcal{I}_\mathcal{T})\\
&=\sum_{g_i\in\mathcal{T}}\mathbf{f}_i^\top L\mathbf{f}_i
\label{eq:J}
\end{split}
\end{equation}
where $\text{tr}(\cdot)$ is the trace of a matrix, $D$ is a diagonal matrix whose entries are column sums of $\Phi$, \emph{i.e.}, $D_{ii}=\sum_{j}\Phi_{ij}$, and $L=D-\Phi$ is a Laplacian matrix.

\begin{Definition}[\gscore]\label{def:q}
Let $\mathcal{D}=\{G_1,\cdots,G_n\}$ denote a graph dataset with multiple side views. Suppose $\Phi$ is a matrix defined as Eq.~(\ref{eq:W}), and $L$ is a Laplacian matrix defined as $L=D-\Phi$, where $D$ is a diagonal matrix, $D_{ii}=\sum_{j}\Phi_{ij}$. We define an evaluation criterion $q$, called {\gscore}, for a subgraph pattern $g_i$ as
\begin{equation}
q(g_i)=\mathbf{f}_i^\top L\mathbf{f}_i
\label{eq:q}
\end{equation}
where $\mathbf{f}_i=[f_{i1},\cdots,f_{in}]^\top\in\{0,1\}^n$ is the indicator vector for subgraph pattern $g_i$, $f_{ij}=1$ iff $g_i\subseteq G_j$, otherwise $f_{ij}=0$. Since the Laplacian matrix $L$ is positive semi-definite, for any subgraph pattern $g_i$, $q(g_i)\ge0$.
\end{Definition}

Based on {\gscore} as defined above, the optimization problem in Eq.~(\ref{eq:argmin1}) can be written as
\begin{equation}
\mathcal{T}^*=\operatornamewithlimits{argmin}_{\mathcal{T}\subseteq\mathcal{S}}\sum_{g_i\in\mathcal{T}}q(g_i)~~~\text{s.t.}~|\mathcal{T}|\le k
\label{eq:argmin2}
\end{equation}

The optimal solution to the problem in Eq.~(\ref{eq:argmin2}) can be found by using {\gscore} to conduct feature selection on a set of subgraph patterns in $\mathcal{S}$. Suppose the {\gscore} values for all subgraph patterns are denoted as $q(g_1)\le\cdots\le q(g_m)$ in sorted order, then the optimal solution to the optimization problem in Eq.~(\ref{eq:argmin2}) is
\begin{equation}
\mathcal{T}^*=\operatornamewithlimits{\cup}_{i=1}^k\{g_i\}
\label{eq:T}
\end{equation}

\subsection{Searching with A Lower Bound: \galgo}
%\label{sec:solution}

Now we address the second problem discussed in Section~\ref{sec:problem}, and propose an efficient method to find the optimal set of subgraph patterns from a graph dataset with multiple side views.

A straightforward solution to the goal of finding an optimal feature set is the exhaustive enumeration, \emph{i.e.}, we could first enumerate all subgraph patterns from a graph dataset, and then calculate the {\gscore} values for all subgraph patterns. In the context of graph data, however, it is usually not feasible to enumerate the full set of subgraph patterns before feature selection. Actually, the number of subgraph patterns grows exponentially with the size of graphs. Inspired by recent advances in graph classification approaches \cite{cao2015identification,kong2010multi,kong2010semi,yan2008mining}, which nest their evaluation criteria into the subgraph mining process and develop constraints to prune the search space, we adopt a similar approach by deriving a different constraint based upon {\gscore}.

By adopting the gSpan algorithm proposed by Yan and Han \cite{yan2002gspan}, we can enumerate all the subgraph patterns for a graph dataset in a canonical search space. In order to prune the subgraph search space, we now derive a lower bound of the {\gscore} value:
\begin{Theorem}\label{theorem:bound}
Given any two subgraph patterns $g_i,g_j\in\mathcal{S}$, $g_j$ is a supergraph of $g_i$, \emph{i.e.}, $g_i\subseteq g_j$. The {\gscore} value of $g_j$ is bounded by $\hat{q}(g_i)$, \emph{i.e.}, $q(g_j)\ge\hat{q}(g_i)$. $\hat{q}(g_i)$ is defined as
\begin{equation}
\hat{q}(g_i)\triangleq\mathbf{f}_i^\top\hat{L}\mathbf{f}_i
\label{eq:bound}
\end{equation}
where the matrix $\hat{L}$ is defined as $\hat{L}_{pq}\triangleq\min(0,L_{pq})$.
\end{Theorem}
\textsc{Proof}.
According to Definition~\ref{def:q},
\begin{equation}
q(g_j)=\mathbf{f}_j^\top L\mathbf{f}_j=\sum_{p,q:G_p,G_q\in\mathcal{G}(g_j)}L_{pq}
\label{eq:proof1}
\end{equation}
where $\mathcal{G}(g_j)\triangleq\{G_k|g_j\subseteq G_k,1\le k\le n\}$. Since $g_i\subseteq g_j$, according to anti-monotonic property, we have $\mathcal{G}(g_j)\subseteq\mathcal{G}(g_i)$. Also $\hat{L}_{pq}\triangleq\min(0,L_{pq})$, we have $\hat{L}_{pq}\le L_{pq}$ and $\hat{L}_{pq}\le0$. Therefore,
\begin{equation}
\begin{split}
q(g_j)&=\sum_{p,q:G_p,G_q\in\mathcal{G}(g_j)}L_{pq}\ge\sum_{p,q:G_p,G_q\in\mathcal{G}(g_j)}\hat{L}_{pq}\\
&\ge\sum_{p,q:G_p,G_q\in\mathcal{G}(g_i)}\hat{L}_{pq}=\hat{q}(g_i)
\label{eq:proof2}
\end{split}
\end{equation}
Thus, for any $g_i\subseteq g_j$, $q(g_j)\ge\hat{q}(g_i)$. $\square$

\begin{algorithm}
\caption{The Proposed Method: \galgo}
\label{alg1}
\begin{algorithmic}[1]
\REQUIRE $\mathcal{D}, min\_sup, k, \{\lambda^{(p)},\kappa^{(p)}\}_{p=1}^{v}$
\ENSURE $\mathcal{T}$: Set of optimal subgraph patterns
\STATE $\mathcal{T}=\emptyset, \theta=Inf$
\WHILE{$\text{unexplored nodes in the DFS code tree}\ne\emptyset$}
\STATE $g=\text{currently explored node in the DFS code tree}$
\IF{$freq(g)\ge min\_sup$}
\IF{$|\mathcal{T}|<k~\text{or}~q(g)<\theta$}
\STATE $\mathcal{T}=\mathcal{T}\cup\{g\}$
\IF{$|\mathcal{T}|>k$}
\STATE $g_{max}=\operatornamewithlimits{argmax}_{g'\in\mathcal{T}}q(g')$
\STATE $\mathcal{T}=\mathcal{T}/\{g_{max}\}$
\ENDIF
\STATE $\theta=\max_{g'\in\mathcal{T}}q(g')$
\ENDIF
\IF{$\hat{q}(g)<\theta$}
\STATE $\text{Depth-first search the subtree rooted from}~g$
\ENDIF
\ENDIF
\ENDWHILE
\STATE $\text{return}~\mathcal{T}$
\end{algorithmic}
\end{algorithm}

%\subsection{Search Space Pruning}

We can now nest the lower bound into the subgraph mining steps in gSpan to efficiently prune the DFS code tree. During the depth-first search through the DFS code tree, we always maintain the currently top-$k$ best subgraph patterns according to {\gscore} and the temporally suboptimal {\gscore} value (denoted by $\theta$) among all the {\gscore} values calculated before. If $\hat{q}(g_i)\ge\theta$, the {\gscore} value of any supergraph $g_j$ of $g_i$ should be no less than $\hat{q}(g_i)$ according to Theorem~\ref{theorem:bound}, \emph{i.e.}, $q(g_j)\ge\hat{q}(g_i)\ge\theta$. Thus, we can safely prune the subtree rooted from $g_i$ in the search space. If $\hat{q}(g_i)<\theta$, we cannot prune this subtree since there might exist a supergraph $g_j$ of $g_i$ such that $q(g_j)<\theta$. As long as a subgraph $g_i$ can improve the {\gscore} values of any subgraphs in $\mathcal{T}$, it is added into $\mathcal{T}$ and the least best subgraph is removed from $\mathcal{T}$. Then we recursively search for the next subgraph in the DFS code tree. The branch-and-bound algorithm {\galgo} is summarized in Algorithm~\ref{alg1}.

\section{Experiments}
\label{sec:exp}

In order to evaluate the performance of the proposed solution to the problem of feature selection for graph classification using multiple side views, we tested our algorithm on brain network datasets derived from neuroimaging, as introduced in Section~\ref{sec:dataset}.

\subsection{Experimental Setup}

To the best of our knowledge, this paper is the first work on leveraging side information in feature selection problem for graph classification. In order to evaluate the performance of the proposed method, we compare our method with other methods using different statistical measures and discriminative score functions. For all the compared methods, gSpan \cite{yan2002gspan} is used as the underlying searching strategy. Note that although alternative algorithms are available \cite{yan2008mining,jin2009graph,jin2010gaia}, the search step efficiency is not the focus of this paper. The compared methods are summarized as follows:
\begin{itemize}
\item {\galgo}: The proposed discriminative subgraph selection method using multiple side views. Following the observation in Section~\ref{sec:ttest} that side information consistency is verified to be significant in all the side views, the parameters in {\galgo} are simply set to $\lambda^{(1)}=\cdots=\lambda^{(v)}=1$ for experimental purposes. In the case where some side views are suspect to be redundant, we can adopt the alternative optimization strategy to iteratively select discriminative subgraph patterns and update view weights.
\item {\gssc}: A semi-supervised feature selection method for graph classification based upon both labeled and unlabeled graphs. The parameters in gSSC are set to $\alpha=\beta=1$ unless otherwise specified \cite{kong2010semi}.
\item Discriminative Subgraphs ({\duga}, {\dugb}, {\dugc}, {\dugd}): Supervised feature selection methods for graph classification based upon confidence \cite{gao2010direct}, frequency ratio \cite{jin2011lts,jin2010gaia,jin2009graph}, G-test score \cite{yan2008mining} and HSIC \cite{kong2010multi}, respectively. The top-k discriminative subgraph features are selected in terms of different discrimination criteria.
\item Frequent Subgraphs ({\topk}): In this approach, the evaluation criterion for subgraph feature selection is based upon frequency. The top-k frequent subgraph features are selected.
\end{itemize}

We append the side view data to the subgraph-based graph representations computed by the above algorithms before feeding the concatenated feature vectors to the classifier. Another baseline that only uses side view data is denoted as {\side}.

For a fair comparison, we used LibSVM \cite{libsvm} with linear kernel as the base classifier for all the compared methods. In the experiments, 3-fold cross validations were performed on balanced datasets. To get the binary links, we performed simple thresholding over the weights of the links. The \emph{threshold} for fMRI and DTI datasets was 0.9 and 0.3, respectively. 
%The minimum support threshold \emph{min\_sup} in the gSpan was set as 30\% for both datasets. 

\subsection{Performance on Graph Classification}

\begin{figure}[t]
\centering
    \begin{minipage}[l]{\columnwidth}
      \centering
      \includegraphics[width=1\textwidth]{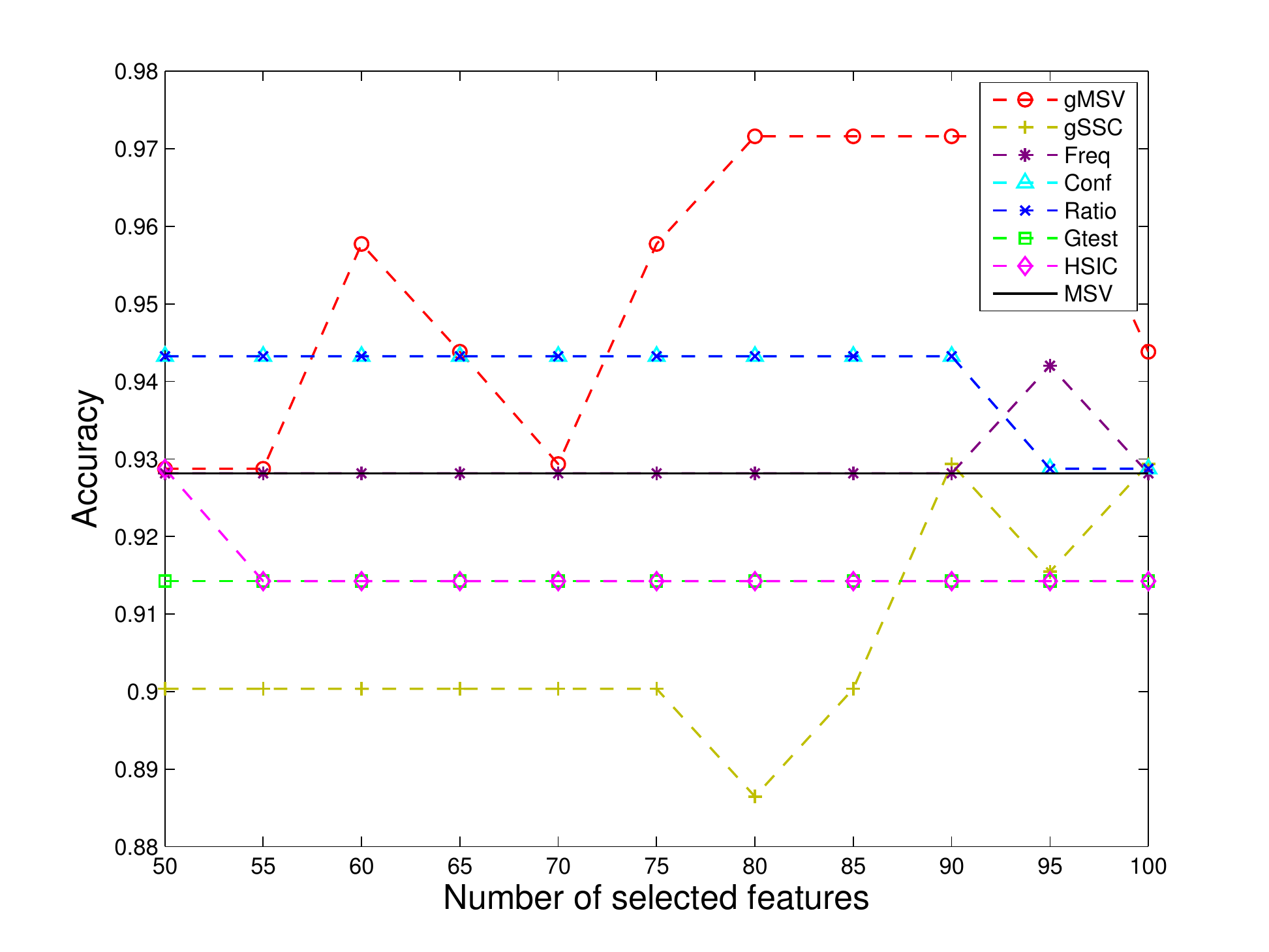}
    \end{minipage}
\caption{Classification performance on the fMRI dataset with different number of features.}\label{fig:fMRI}
\end{figure}

\begin{figure}[t]
\centering
    \begin{minipage}[l]{\columnwidth}
      \centering
      \includegraphics[width=1\textwidth]{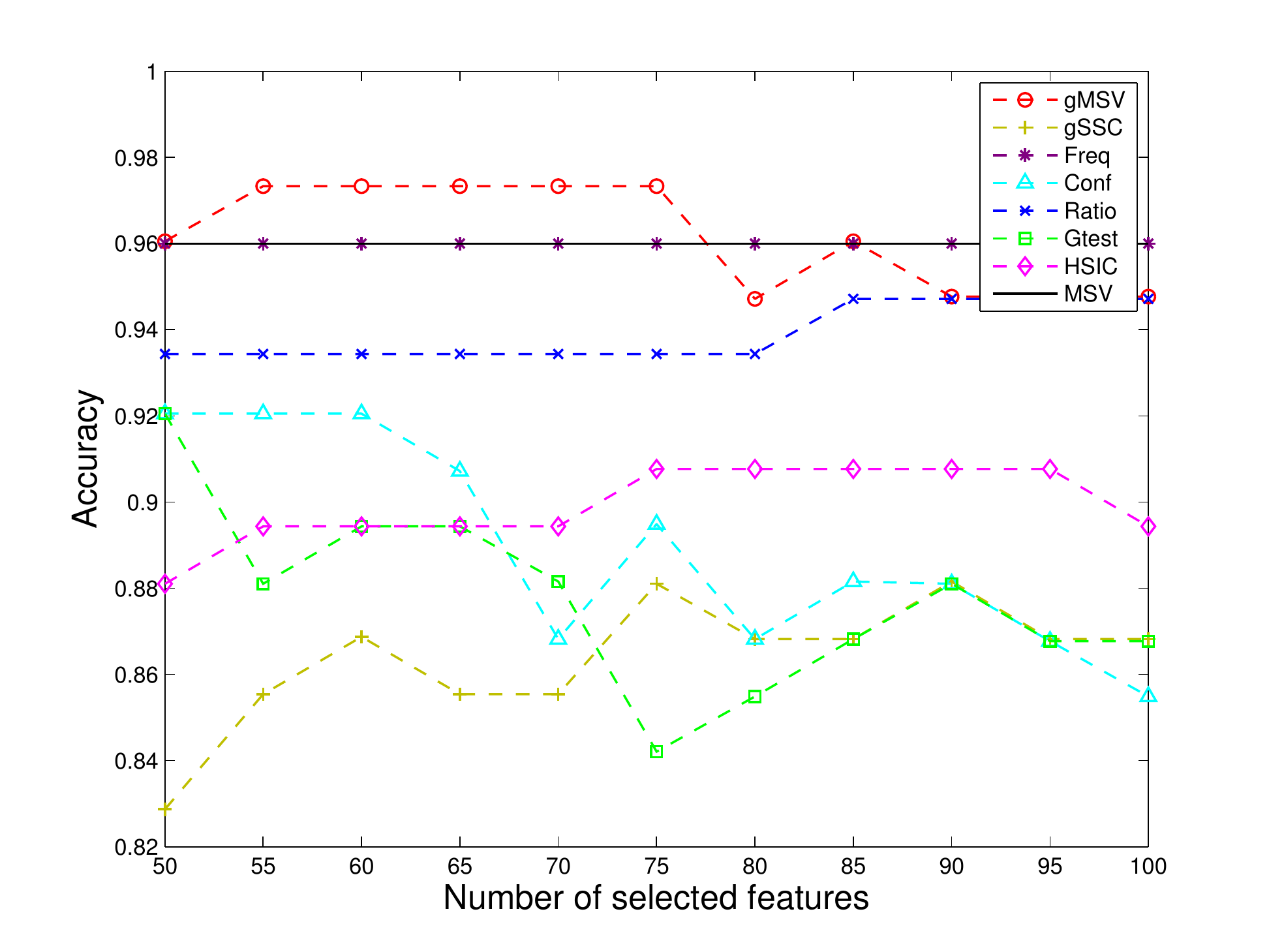}
    \end{minipage}
\caption{Classification performance on the DTI dataset with different number of features.}\label{fig:DTI}
\end{figure}

The experimental results on fMRI and DTI datasets are shown in Figure~\ref{fig:fMRI} and Figure~\ref{fig:DTI}, respectively. The average performances with different number of features of each method are reported. Classification accuracy is used as the evaluation metric.

In Figure~\ref{fig:fMRI}, our method {\galgo} can achieve the classification accuracy as high as 97.16\% on the fMRI dataset, which is significantly better than the union of other subgraph-based features and side view features. The black solid line denotes the method {\side}, the simplest baseline that uses only side view data. {\duga} and {\dugb} can do slightly better than {\side}. {\topk} adopts an unsupervised process for selecting subgraph patterns, resulting in a comparable performance with {\side}, indicating that there is no additional information from the selected subgraphs. Other methods that use different discrimination scores without leveraging the guidance from side views perform even worse than {\side} in graph classification, because they evaluate the usefulness of subgraph patterns solely based on the limited label information from a small sample size of brain networks. The selected subgraph patterns can potentially be redundant or irrelevant, thereby compromising the effects of side view data. Importantly, {\galgo} outperforms the semi-supervised approach {\gssc} which explores the unlabeled graphs based on the separability property. This indicates that rather than simply considering that unlabeled graphs should be separated from each other, it would be better to regularize such separability/closeness to be consistent with the available side views.
 
Similar observations can be found in Figure~\ref{fig:DTI}, where {\galgo} outperforms other baselines by achieving a good performance as high as 97.33\% accuracy on the DTI dataset. We notice that only {\galgo} is able to do better than {\side} by adding complementary subgraph-based features to the side view features. Moreover, the performances of other schemes are not consistent over the two datasets. The 2nd and 3rd best schemes, {\duga} and {\dugb}, for fMRI do not perform as well for DTI. These results support our premise that exploring a plurality of side views can boost the performance of graph classification, and the {\gscore} evaluation criterion in {\galgo} can find more informative subgraph patterns for graph classification than subgraphs based on frequency or other discrimination scores.

\subsection{Time and Space Complexity}

Next, we evaluate the effectiveness of pruning the subgraph search space by adopting the lower bound of {\gscore} in {\galgo}. In this section, we compare the runtime performance of two implementation versions of {\galgo}: the pruning {\galgo} uses the lower bound of {\gscore} to prune the search space of subgraph enumerations, as shown in Algorithm~\ref{alg1}; the unpruning {\galgo} denotes the method without pruning in the subgraph mining process, \emph{e.g.}, deleting the line 13 in Algorithm~\ref{alg1}. We test both approaches and recorded the average CPU time used and the average number of subgraph patterns explored during the procedure of subgraph mining and feature selection.

\begin{figure}
\centering
  \subfigure[fMRI dataset]{\label{fig:fMRI_time}
    \begin{minipage}[l]{0.45\columnwidth}
      \centering
      \includegraphics[width=1\textwidth]{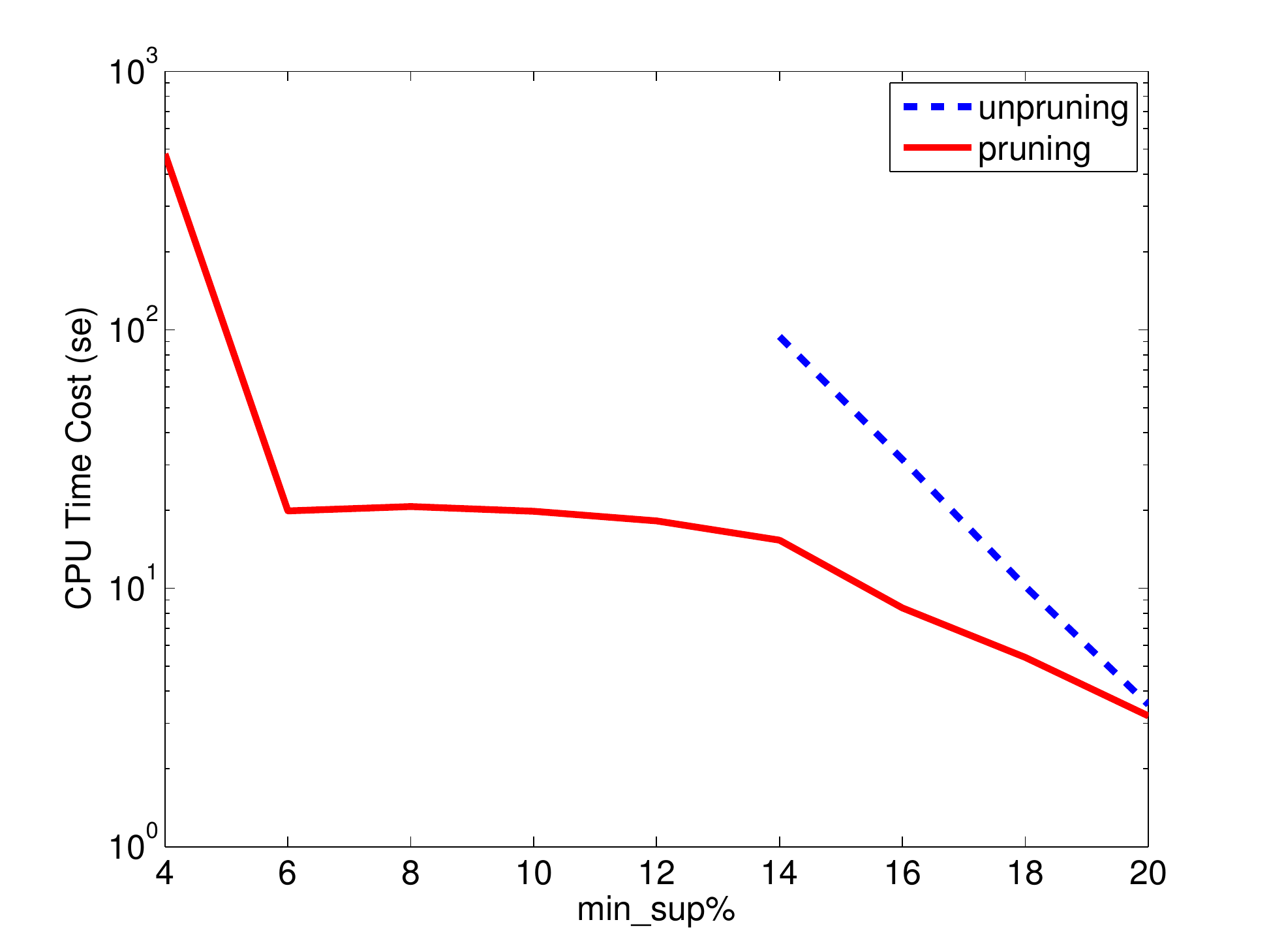}
    \end{minipage}
  }
\subfigure[DTI dataset]{\label{fig:DTI_time}
    \begin{minipage}[l]{0.45\columnwidth}
      \centering
      \includegraphics[width=1\textwidth]{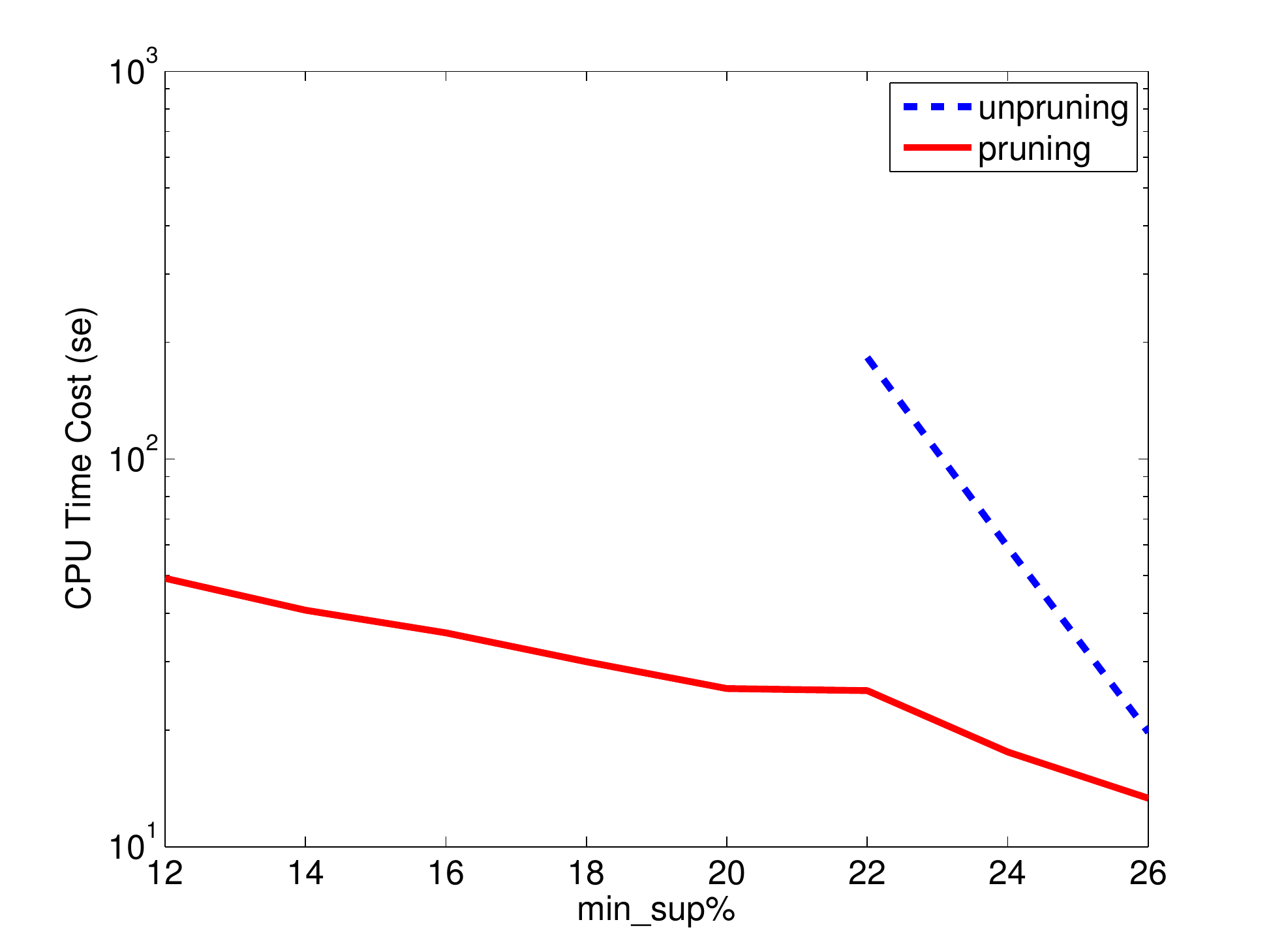}
    \end{minipage}
  }
\caption{Average CPU time for pruning versus unpruning with varying min\_sup.}\label{fig:time}
\end{figure}

\begin{figure}
\centering
  \subfigure[fMRI dataset]{\label{fig:fMRI_num_fea}
    \begin{minipage}[l]{0.45\columnwidth}
      \centering
      \includegraphics[width=1\textwidth]{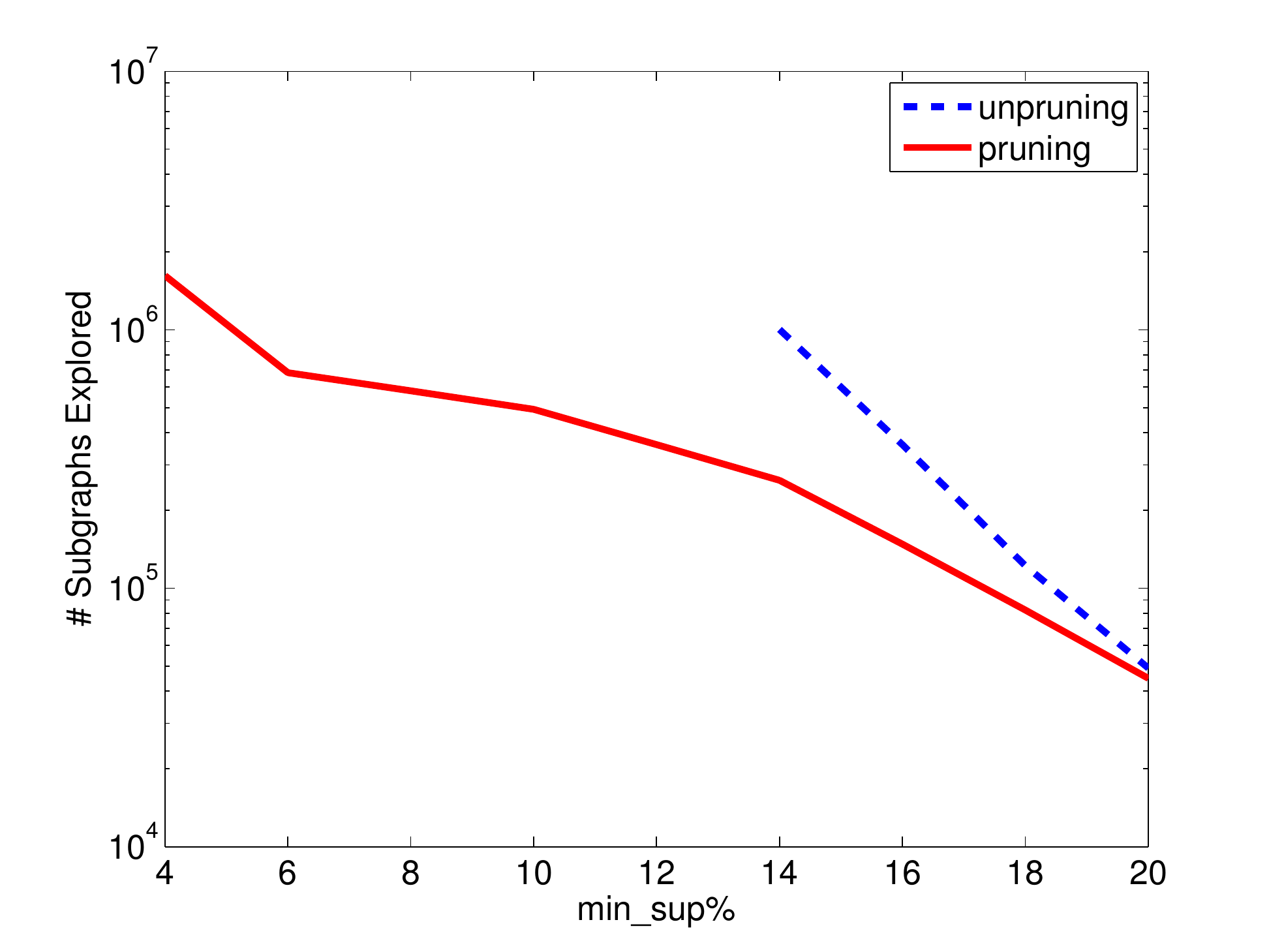}
    \end{minipage}
  }
\subfigure[DTI dataset]{\label{fig:DTI_num_fea}
    \begin{minipage}[l]{0.45\columnwidth}
      \centering
      \includegraphics[width=1\textwidth]{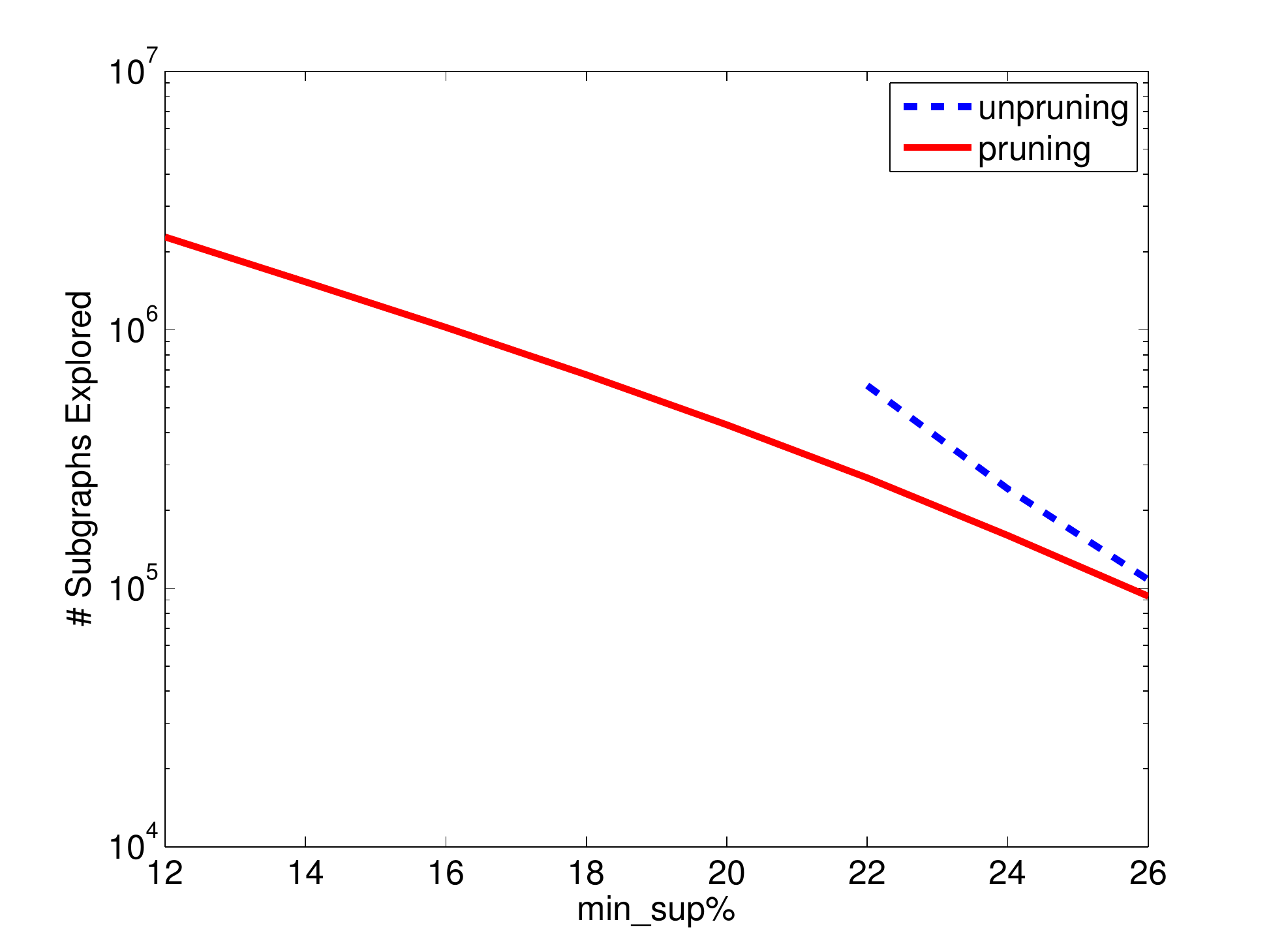}
    \end{minipage}
  }
\caption{Average number of subgraph patterns explored in the mining procedure for pruning versus unpruning with varying min\_sup.}\label{fig:num_fea}
\end{figure}

The comparisons with respect to the time complexity and the space complexity are shown in Figure~\ref{fig:time} and Figure~\ref{fig:num_fea}, respectively. On both datasets, the unpruning {\galgo} needs to explore exponentially larger subgraph search space as we decrease the \emph{min\_sup} value in the subgraph mining process. When the \emph{min\_sup} value is too low, the subgraph enumeration step in the unpruning {\galgo} can run out of the memory. However, the pruning {\galgo} is still effective and efficient when the \emph{min\_sup} value goes to very low, because its running time and space requirement do not increase as much as the unpruning {\galgo} by reducing the subgraph search space via the lower bound of {\gscore}.

The focus of this paper is to investigate side information consistency and explore multiple side views in discriminative subgraph selection. As potential alternatives to the gSpan-based branch-and-bound algorithm, we could employ other more sophisticated searching strategies with our proposed multi-side-view evaluation criterion, {\gscore}. For example, we can replace with {\gscore} the G-test score in LEAP \cite{yan2008mining} or the log ratio in COM \cite{jin2009graph} and GAIA \cite{jin2010gaia}, \emph{etc.} However, as shown in Figure~\ref{fig:time} and Figure~\ref{fig:num_fea}, our proposed solution with pruning, {\galgo}, can survive at $min\_sup=4\%$; considering the limited number of subjects in medical experiments as introduced in Section~\ref{sec:dataset}, {\galgo} is efficient enough for neurological disorder identification where subgraph patterns with too few supported graphs are not desired.

\subsection{Effects of Side Views}

In this section, we first investigate the different contributions from different side views. Table~\ref{tab:result3} shows the performance of {\galgo} on the fMRI dataset by considering only one side view each time. In general, the best performance is achieved by simultaneously exploring all the side views. Specifically, we observe that the side view {\flo} can independently provide the most informative side information for selecting discriminative subgraph patterns on the fMRI brain networks, which might imply that HIV brain injuries in the sense of functional connectivity are most likely to express in measurements from this side view. It is consistent with our finding in Section~\ref{sec:ttest} that the side view {\flo} is the most significantly correlated with the prespecified label information. Results on the DTI dataset are shown in Table~\ref{tab:result4}.

\begin{table}
\caption{Average classification performances of {\galgo} on the fMRI dataset with different single side views.}
\small
\label{tab:result3}
\centering
\begin{tabular}{lcccc}
\toprule%----------------------------
%& \multicolumn{4}{c}{Evaluations}\\
%\cmidrule{2-5}%----------------------
Side views	&Acc. &Prec. &Rec. &F1\\
\midrule %---------------------------
{\neu}	&0.743 &0.851 &0.679 &0.734\\
{\flo}	&0.887 &0.919 &0.872 &0.892\\
{\pla}	&0.715 &0.769 &0.682 &0.710\\
{\fre}	&0.786 &0.851 &0.737 &0.785\\
{\ave}	&0.672 &0.824 &0.500 &0.618\\
{\dti}	&0.628 &0.686 &0.605 &0.637\\
{\seg}	&0.701 &0.739 &0.737 &0.731\\
\midrule %---------------------------
All side views	&0.972 &1.000 &0.949 &0.973\\
\bottomrule%-------------------------
\end{tabular}
\end{table}

\begin{table}
\caption{Average classification performances of {\galgo} on the DTI dataset with different single side views.}
\small
\label{tab:result4}
\centering
\begin{tabular}{lcccc}
\toprule%----------------------------
%& \multicolumn{4}{c}{Evaluations}\\
%\cmidrule{2-5}%----------------------
Side views	&Acc. &Prec. &Rec. &F1\\
\midrule %---------------------------
{\neu}	&0.616 &0.630 &0.705 &0.662\\
{\flo}	&0.815 &0.847 &0.808 &0.822\\
{\pla}	&0.736 &0.801 &0.705 &0.744\\
{\fre}	&0.631 &0.664 &0.632 &0.644\\
{\ave}	&0.604 &0.626 &0.679 &0.647\\
{\dti}	&0.723 &0.717 &0.775 &0.741\\
{\seg}	&0.605 &0.616 &0.679 &0.644\\
\midrule %---------------------------
All side views	&0.973 &1.000 &0.951 &0.974\\
\bottomrule%-------------------------
\end{tabular}
\end{table}

\subsection{Feature Evaluation}

\begin{figure}[!ht]
\centering
  \subfigure[]{\label{fig:fMRI_1}
    \begin{minipage}[l]{0.45\columnwidth}
      \centering
      \includegraphics[width=1\textwidth]{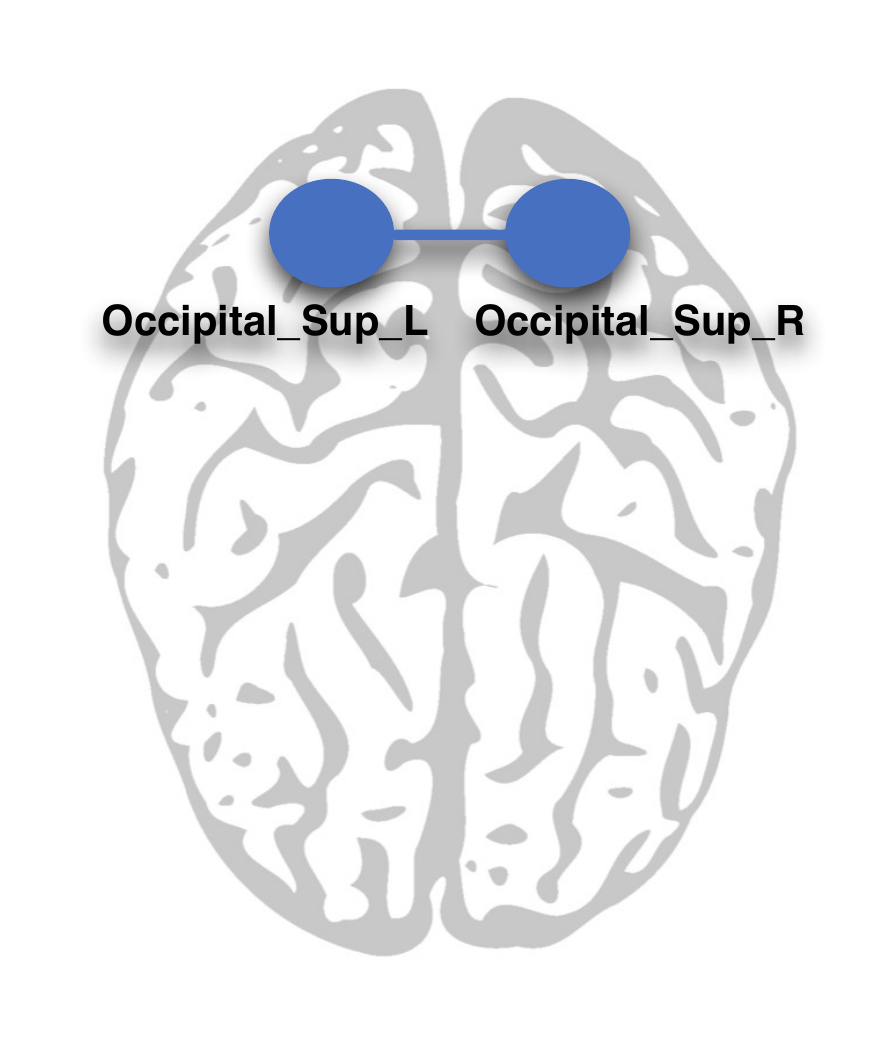}
    \end{minipage}
  }
  \subfigure[]{\label{fig:fMRI_2}
    \begin{minipage}[l]{0.45\columnwidth}
      \centering
      \includegraphics[width=1\textwidth]{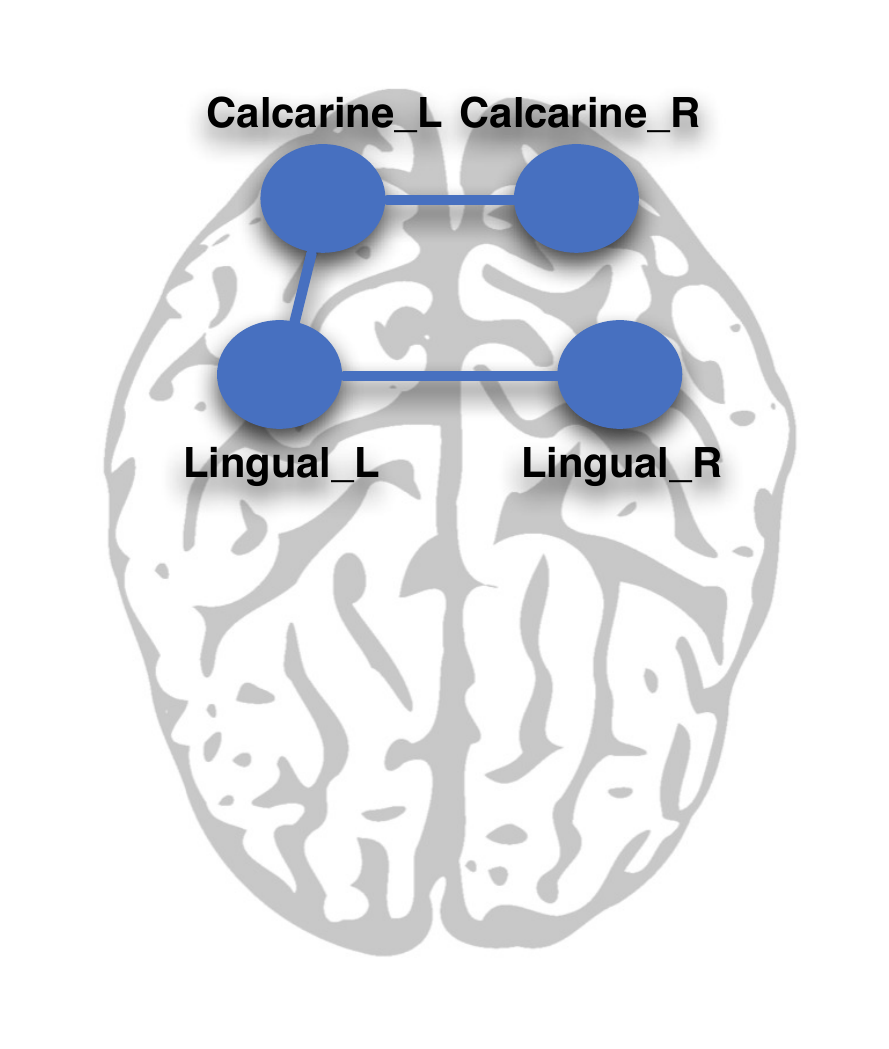}
    \end{minipage}
  }
\caption{Discriminative subgraph patterns that are associated with HIV, selected from the fMRI dataset.}\label{fig:fMRI_fea}
\end{figure}

\begin{figure}[!ht]
\centering
  \subfigure[]{\label{fig:DTI_1}
    \begin{minipage}[l]{0.45\columnwidth}
      \centering
      \includegraphics[width=1\textwidth]{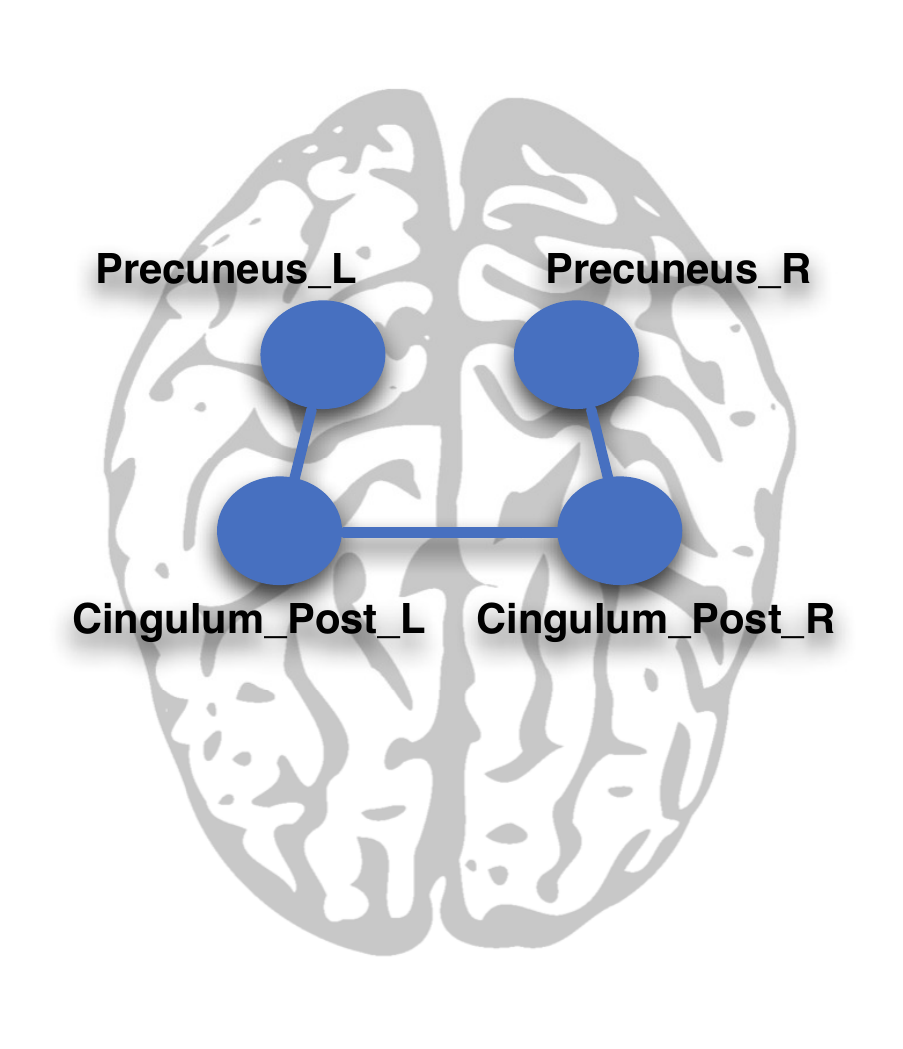}
    \end{minipage}
  }
  \subfigure[]{\label{fig:DTI_2}
    \begin{minipage}[l]{0.45\columnwidth}
      \centering
      \includegraphics[width=1\textwidth]{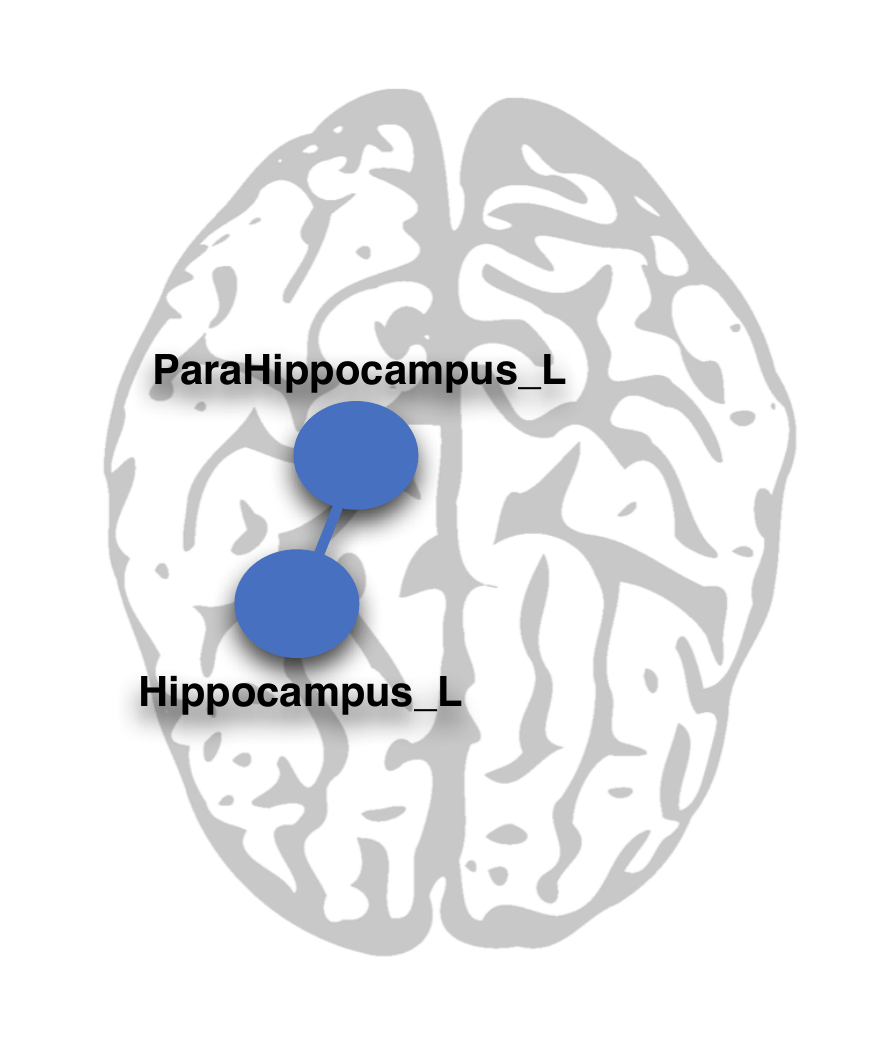}
    \end{minipage}
  }
\caption{Discriminative subgraph patterns that are associated with HIV, selected from the DTI dataset.}\label{fig:DTI_fea}
\end{figure}

Figure~\ref{fig:fMRI_fea} and Figure~\ref{fig:DTI_fea} display the most discriminative subgraph patterns selected by {\galgo} from the fMRI dataset and the DTI dataset, respectively. These findings examining functional and structural networks are consistent with other in vivo studies \cite{castelo2006altered,wang2011abnormalities} and with the pattern of brain injury at autopsy \cite{everall1993neuronal,langford2003changing} in HIV infection. With the approach presented in this analysis, alterations in the brain can be detected in initial stages of injury and in the context of clinically meaningful information, such as host immune status and immune response ({\flo}), immune mediators ({\pla}) and cognitive function ({\neu}). This approach optimizes the valuable information inherent in complex clinical datasets. Strategies for combining various sources of clinical information have promising potential for informing an understanding of  disease mechanisms, for identification of new therapeutic targets and for discovery of biomarkers to assess risk and to evaluate response to treatment.

%-----------------------------------------------
\section{Related Work}
\label{sec:relatedwork}

To the best of our knowledge, this paper is the first work exploring side information in the task of subgraph feature selection for graph classification. Our work is related to subgraph mining techniques and multi-view feature selection problems. We briefly discuss both of them.

Mining subgraph patterns from graph data has been studied extensively by many researchers. In general, a variety of filtering criteria are proposed. A typical evaluation criterion is frequency, which aims at searching for frequently appearing subgraph features in a graph dataset satisfying a prespecified \emph{min\_sup} value. Most of the frequent subgraph mining approaches are unsupervised. For example, Yan and Han developed a depth-first search algorithm: gSpan \cite{yan2002gspan}. This algorithm builds a lexicographic order among graphs, and maps each graph to an unique minimum DFS code as its canonical label. Based on this lexicographic order, gSpan adopts the depth-first search strategy to mine frequent connected subgraphs efficiently. Many other approaches for frequent subgraph mining have also been proposed, \emph{e.g.}, AGM \cite{inokuchi2000apriori}, FSG \cite{kuramochi2001frequent}, MoFa \cite{borgelt2002mining}, FFSM \cite{huan2003efficient}, and Gaston \cite{nijssen2004quickstart}.

Moreover, the problem of supervised subgraph mining has been studied in recent work which examines how to improve the efficiency of searching the discriminative subgraph patterns for graph classification. Yan et al. introduced two concepts \emph{structural leap search} and \emph{frequency-descending mining}, and proposed LEAP \cite{yan2008mining} which is one of the first work in discriminative subgraph mining. Thoma et al. proposed CORK which can yield a near-optimal solution using greedy feature selection \cite{thoma2009near}. Ranu and Singh proposed a scalable approach, called GraphSig, that is capable of mining discriminative subgraphs with a low frequency threshold \cite{ranu2009graphsig}. Jin et al. proposed COM which takes into account the co-occurences of subgraph patterns, thereby facilitating the mining process \cite{jin2009graph}. Jin et al. further proposed an evolutionary computation  method, called GAIA, to mine discriminative subgraph patterns using a randomized searching strategy \cite{jin2010gaia}. Our proposed criterion {\gscore} can be combined with these efficient searching algorithms to speed up the process of mining discriminative subgraph patterns by substituting the G-test score in LEAP \cite{yan2008mining} or the log ratio in COM \cite{jin2009graph} and GAIA \cite{jin2010gaia}, \emph{etc.} Zhu et al. designed a diversified discrimination score based on the log ratio which can reduce the overlap between selected features by considering the embedding overlaps in the graphs \cite{zhu2012graph}. Similar idea can be integrated into {\gscore} to improve feature diversity.

There are some recent works on incorporating multi-view learning and feature selection. Tang et al. studied unsupervised multi-view feature selection by constraining that similar data instances from each view should have similar pseudo-class labels \cite{tang2013unsupervised}. Cao et al. explored tensor product to bring different views together in a joint space and presents a dual method of tensor-based multi-view feature selection \cite{cao2014tensor}. Aggarwal et al. considered side information for text mining \cite{aggarwal2012use}. However, these methods are limited in requiring a set of candidate features as input, and therefore are not directly applicable for graph data. Wu et al. considered the scenario where one object can be described by multiple graphs generated from different feature views and proposes an evaluation criterion to estimate the discriminative power and the redundancy of subgraph features across all views \cite{wu2014multi}. In contrast, in this paper, we assume that one object can have other data representations of side views in addition to the primary graph view.

In the context of graph data, the subgraph features are embedded within the complex graph structures and usually it is not feasible to enumerate the full set of features for a graph dataset before the feature selection. Actually, the number of subgraph features grows exponentially with the size of graphs. In this paper, we explore the side information from multiple views to effectively facilitate the procedure of discriminative subgraph mining. Our proposed feature selection for graph data is integrated to the subgraph mining process, which can efficiently prune the search space, thereby avoiding exhaustive enumeration of all subgraph features.

%-----------------------------------------------
\section{Conclusion and Future Work}
\label{sec:conclusion}

We presented an approach for selecting discriminative subgraph features using multiple side views. This has important applications in neurological disorder diagnosis via brain networks. We show in this paper that by leveraging the information from multiple side views that are available along with the graph data, the proposed method {\galgo} can achieve very good performance on the problem of feature selection for graph classification, and the selected subgraph patterns are relevant to disease diagnosis.

A potential extension to our method is to combine fMRI and DTI brain networks to find discriminative subgraph patterns in the sense of both functional and structural connections. Other extensions include better exploring weighted links in the multi-side-view setting. It is also interesting to have our model applied to other domains where one can find graph data and side information aligned with the graph. For example, in bioinformatics, chemical compounds can be represented by graphs based on their inherent molecular structures and are associated with properties such as drug repositioning, side effects, ontology annotations. Leveraging all these information to find out discriminative subgraph patterns can be transformative for drug discovery.

\section{Acknowledgements}
This work is supported in part by NSF through grants III-1526499, CNS-1115234, and OISE-1129076, Google Research Award, the Pinnacle Lab at Singapore Management University, and NIH through grant R01-MH080636.

%-----------------------------------------------
\balance
\bibliographystyle{plain}
\bibliography{reference}

\begin{thebibliography}{10}

\bibitem{aggarwal2012use}
Charu~C Aggarwal, Yuchen Zhao, and Philip~S Yu.
\newblock On the use of side information for mining text data.
\newblock {\em TKDE}, pages 1--1, 2012.

\bibitem{bar2005learning}
Aharon Bar-Hillel, Tomer Hertz, Noam Shental, and Daphna Weinshall.
\newblock Learning a mahalanobis metric from equivalence constraints.
\newblock {\em Journal of Machine Learning Research}, 6(6):937--965, 2005.

\bibitem{borgelt2002mining}
Christian Borgelt and Michael~R Berthold.
\newblock Mining molecular fragments: Finding relevant substructures of
  molecules.
\newblock In {\em ICDM}, pages 51--58. IEEE, 2002.

\bibitem{cao2014tensor}
Bokai Cao, Lifang He, Xiangnan Kong, Philip~S Yu, Zhifeng Hao, and Ann~B Ragin.
\newblock Tensor-based multi-view feature selection with applications to brain
  diseases.
\newblock In {\em ICDM}, pages 40--49. IEEE, 2014.

\bibitem{cao2015determinants}
Bokai Cao, Xiangnan Kong, Casey Kettering, Philip~S. Yu, and Ann~B. Ragin.
\newblock Determinants of {HIV}-induced brain changes in three different
  periods of the early clinical course: A data mining analysis.
\newblock {\em NeuroImage: Clinical}, 2015.

\bibitem{cao2015identification}
Bokai Cao, Liang Zhan, Xiangnan Kong, Philip~S. Yu, Nathalie Vizueta, Lori~L.
  Altshuler, and Alex~D. Leow.
\newblock Identification of discriminative subgraph patterns in {fMRI} brain
  networks in bipolar affective disorder.
\newblock In {\em Brain Informatics and Health}. Springer, 2015.

\bibitem{castelo2006altered}
JMB Castelo, SJ~Sherman, MG~Courtney, RJ~Melrose, and CE~Stern.
\newblock Altered hippocampal-prefrontal activation in {HIV} patients during
  episodic memory encoding.
\newblock {\em Neurology}, 66(11):1688--1695, 2006.

\bibitem{libsvm}
Chih-Chung Chang and Chih-Jen Lin.
\newblock {\em {LIBSVM}: a library for support vector machines}, 2001.
\newblock Software available at \url{http://www.csie.ntu.edu.tw/~cjlin/libsvm}.

\bibitem{dai2007co}
Wenyuan Dai, Gui-Rong Xue, Qiang Yang, and Yong Yu.
\newblock Co-clustering based classification for out-of-domain documents.
\newblock In {\em KDD}, pages 210--219. ACM, 2007.

\bibitem{everall1993neuronal}
Ian~Paul Everall, Philip~J Luthert, and Peter~L Lantos.
\newblock Neuronal number and volume alterations in the neocortex of hiv
  infected individuals.
\newblock {\em Journal of Neurology, Neurosurgery \& Psychiatry},
  56(5):481--486, 1993.

\bibitem{gao2010direct}
Chuancong Gao and Jianyong Wang.
\newblock Direct mining of discriminative patterns for classifying uncertain
  data.
\newblock In {\em KDD}, pages 861--870. ACM, 2010.

\bibitem{huan2003efficient}
Jun Huan, Wei Wang, and Jan Prins.
\newblock Efficient mining of frequent subgraphs in the presence of
  isomorphism.
\newblock In {\em ICDM}, pages 549--552. IEEE, 2003.

\bibitem{inokuchi2000apriori}
Akihiro Inokuchi, Takashi Washio, and Hiroshi Motoda.
\newblock An apriori-based algorithm for mining frequent substructures from
  graph data.
\newblock In {\em Principles of Data Mining and Knowledge Discovery}, pages
  13--23. Springer, 2000.

\bibitem{jenkinson2005bet2}
Mark Jenkinson, Mickael Pechaud, and Stephen Smith.
\newblock {BET2}: {MR}-based estimation of brain, skull and scalp surfaces.
\newblock In {\em Eleventh annual meeting of the organization for human brain
  mapping}, volume~17, 2005.

\bibitem{jin2011lts}
Ning Jin and Wei Wang.
\newblock {LTS}: Discriminative subgraph mining by learning from search
  history.
\newblock In {\em ICDE}, pages 207--218. IEEE, 2011.

\bibitem{jin2009graph}
Ning Jin, Calvin Young, and Wei Wang.
\newblock Graph classification based on pattern co-occurrence.
\newblock In {\em CIKM}, pages 573--582. ACM, 2009.

\bibitem{jin2010gaia}
Ning Jin, Calvin Young, and Wei Wang.
\newblock {GAIA}: graph classification using evolutionary computation.
\newblock In {\em SIGMOD}, pages 879--890. ACM, 2010.

\bibitem{kong2013discriminative}
Xiangnan Kong, Ann~B Ragin, Xue Wang, and Philip~S Yu.
\newblock Discriminative feature selection for uncertain graph classification.
\newblock In {\em SDM}, pages 82--93. SIAM, 2013.

\bibitem{kong2010multi}
Xiangnan Kong and Philip~S Yu.
\newblock Multi-label feature selection for graph classification.
\newblock In {\em ICDM}, pages 274--283. IEEE, 2010.

\bibitem{kong2010semi}
Xiangnan Kong and Philip~S Yu.
\newblock Semi-supervised feature selection for graph classification.
\newblock In {\em KDD}, pages 793--802. ACM, 2010.

\bibitem{kuramochi2001frequent}
Michihiro Kuramochi and George Karypis.
\newblock Frequent subgraph discovery.
\newblock In {\em ICDM}, pages 313--320. IEEE, 2001.

\bibitem{langford2003changing}
TD~Langford, SL~Letendre, GJ~Larrea, and E~Masliah.
\newblock Changing patterns in the neuropathogenesis of hiv during the haart
  era.
\newblock {\em Brain pathology}, 13(2):195--210, 2003.

\bibitem{mihalkova2007mapping}
Lilyana Mihalkova, Tuyen Huynh, and Raymond~J Mooney.
\newblock Mapping and revising markov logic networks for transfer learning.
\newblock In {\em AAAI}, volume~7, pages 608--614, 2007.

\bibitem{mihalkova2009transfer}
Lilyana Mihalkova and Raymond~J Mooney.
\newblock Transfer learning from minimal target data by mapping across
  relational domains.
\newblock In {\em IJCAI}, volume~9, pages 1163--1168, 2009.

\bibitem{nijssen2004quickstart}
Siegfried Nijssen and Joost~N Kok.
\newblock A quickstart in frequent structure mining can make a difference.
\newblock In {\em KDD}, pages 647--652. ACM, 2004.

\bibitem{ragin2012structural}
Ann~B Ragin, Hongyan Du, Renee Ochs, Ying Wu, Christina~L Sammet, Alfred
  Shoukry, and Leon~G Epstein.
\newblock Structural brain alterations can be detected early in {HIV}
  infection.
\newblock {\em Neurology}, 79(24):2328--2334, 2012.

\bibitem{ranu2009graphsig}
Sayan Ranu and Ambuj~K Singh.
\newblock Graphsig: A scalable approach to mining significant subgraphs in
  large graph databases.
\newblock In {\em ICDE}, pages 844--855. IEEE, 2009.

\bibitem{shi2012transfer}
Xiaoxiao Shi, Xiangnan Kong, and Philip~S Yu.
\newblock Transfer significant subgraphs across graph databases.
\newblock In {\em SDM}, pages 552--563. SIAM, 2012.

\bibitem{smith2002fast}
Stephen~M Smith.
\newblock Fast robust automated brain extraction.
\newblock {\em Human brain mapping}, 17(3):143--155, 2002.

\bibitem{tang2013unsupervised}
Jiliang Tang, Xia Hu, Huiji Gao, and Huan Liu.
\newblock Unsupervised feature selection for multi-view data in social media.
\newblock In {\em SDM}, pages 270--278. SIAM, 2013.

\bibitem{tang2006pairwise}
Wei Tang and Shi Zhong.
\newblock Pairwise constraints-guided dimensionality reduction.
\newblock In {\em SDM Workshop on Feature Selection for Data Mining}, 2006.

\bibitem{thoma2009near}
Marisa Thoma, Hong Cheng, Arthur Gretton, Jiawei Han, Hans-Peter Kriegel,
  Alexander~J Smola, Le~Song, Philip~S Yu, Xifeng Yan, and Karsten~M Borgwardt.
\newblock Near-optimal supervised feature selection among frequent subgraphs.
\newblock In {\em SDM}, pages 1076--1087. SIAM, 2009.

\bibitem{tzourio2002automated}
Nathalie Tzourio-Mazoyer, Brigitte Landeau, Dimitri Papathanassiou, Fabrice
  Crivello, Olivier Etard, Nicolas Delcroix, Bernard Mazoyer, and Marc Joliot.
\newblock Automated anatomical labeling of activations in {SPM} using a
  macroscopic anatomical parcellation of the {MNI} {MRI} single-subject brain.
\newblock {\em Neuroimage}, 15(1):273--289, 2002.

\bibitem{wang2011abnormalities}
Xue Wang, Paul Foryt, Renee Ochs, Jae-Hoon Chung, Ying Wu, Todd Parrish, and
  Ann~B Ragin.
\newblock Abnormalities in resting-state functional connectivity in early human
  immunodeficiency virus infection.
\newblock {\em Brain connectivity}, 1(3):207--217, 2011.

\bibitem{wu2014multi}
Jia Wu, Zhibin Hong, Shirui Pan, Xingquan Zhu, Zhihua Cai, and Chengqi Zhang.
\newblock Multi-graph-view learning for graph classification.
\newblock In {\em ICDM}, pages 590--599. IEEE, 2014.

\bibitem{yan2008mining}
Xifeng Yan, Hong Cheng, Jiawei Han, and Philip~S Yu.
\newblock Mining significant graph patterns by leap search.
\newblock In {\em SIGMOD}, pages 433--444. ACM, 2008.

\bibitem{yan2002gspan}
Xifeng Yan and Jiawei Han.
\newblock gspan: Graph-based substructure pattern mining.
\newblock In {\em ICDM}, pages 721--724. IEEE, 2002.

\bibitem{zhu2012graph}
Yuanyuan Zhu, Jeffrey~Xu Yu, Hong Cheng, and Lu~Qin.
\newblock Graph classification: a diversified discriminative feature selection
  approach.
\newblock In {\em CIKM}, pages 205--214. ACM, 2012.

\end{thebibliography}

\end{document}